\documentclass[11pt]{article}
\usepackage[final]{acl}
\usepackage{times}
\usepackage{latexsym}
\usepackage{algorithm}
\usepackage{algpseudocode}  % <-- CHANGED from algorithmic to algpseudocode
\usepackage{booktabs, multirow}
\usepackage[T1]{fontenc}
\usepackage{graphicx, subcaption}
\usepackage{amsmath}
\usepackage{listings}
\usepackage{xcolor}

\definecolor{codegreen}{rgb}{0,0.6,0}
\definecolor{codegray}{rgb}{0.5,0.5,0.5}
\definecolor{codepurple}{rgb}{0.58,0,0.82}
\definecolor{backcolour}{rgb}{0.98,0.98,0.98}
\definecolor{keywordblue}{rgb}{0.13,0.13,1}

\lstdefinestyle{pythonstyle}{
    backgroundcolor=\color{backcolour},   
    commentstyle=\color{codegreen},
    keywordstyle=\color{keywordblue}\bfseries,
    numberstyle=\tiny\color{codegray},
    stringstyle=\color{codepurple},
    basicstyle=\ttfamily\scriptsize,
    breakatwhitespace=false,         
    breaklines=true,                 
    captionpos=b,                    
    keepspaces=true,                 
    numbers=left,                    
    numbersep=5pt,                  
    showspaces=false,                
    showstringspaces=false,
    showtabs=false,                  
    tabsize=2,
    frame=single,
    rulecolor=\color{codegray},
    columns=flexible,
    morekeywords={self, True, False, None, def, class, return, for, in, if, else, elif, try, except, with, as, from, import, List, Dict, Any, Optional, Set, Tuple, dataclass}
}

\usepackage[utf8]{inputenc}
\usepackage{microtype}
\usepackage{inconsolata}
\usepackage{graphicx}
\usepackage{amssymb}
\usepackage{amsmath}        % <-- ADD: for math environments
\usepackage{xcolor}         % <-- ADD: for colored comments
\usepackage{colortbl}
\usepackage{tabularx, colortbl}
\usepackage{hyperref}
\usepackage{tikz}
\usetikzlibrary{shapes, arrows.meta, positioning, fit, backgrounds, calc}
\definecolor{commentblue}{RGB}{0,0,180}

\algrenewcommand{\algorithmiccomment}[1]{\hfill{\color{commentblue}$\triangleright$ #1}}

\algtext*{EndFor}
\algtext*{EndIf}
\algtext*{EndFunction}

\makeatletter
\renewcommand{\ALG@beginalgorithmic}{\small}
\makeatother
\newcommand{\smark}[1]{\textsubscript{\textcolor{gray}{\scriptsize #1}}}
\title{Learn Like Humans: Use Meta-cognitive Reflection for Efficient Self-Improvement}

\author{
Xinmeng Hou$^1$, Peiliang Gong$^1$, Bohao Qu$^2$, Wuqi Wang$^3$, Qing Guo$^4$, Yang Liu$^1$ \\
$^1$Nanyang Technological University, Singapore \\
$^2$A*STAR, Singapore \\
$^3$Chang'an University, China \\
$^4$Nankai University, China \\
\texttt{\{hou\_xinmeng, cs-peiliang.gong, yangliu\}@ntu.edu.sg} \\
\texttt{qubohao@126.com, wuqiwang@chd.edu.cn, tsingqguo@ieee.org}
}

\begin{document}
\maketitle
\begin{abstract}
While Large Language Models (LLMs) enable complex autonomous behavior, current agents remain constrained by static, human-designed prompts that limit adaptability. Existing self-improving frameworks attempt to bridge this gap but typically rely on inefficient, multi-turn recursive loops that incur high computational costs. To address this, we propose \textbf{\underline{M}}etacognitive \textbf{\underline{A}}gent \textbf{\underline{R}}eflective \textbf{\underline{S}}elf-improvement (MARS), a framework that achieves efficient self-evolution within a single recurrence cycle. Inspired by educational psychology, MARS mimics human learning by integrating principle-based reflection (abstracting normative rules to avoid errors) and procedural reflection (deriving step-by-step strategies for success). By synthesizing these insights into optimized instructions, MARS allows agents to systematically refine their reasoning logic without continuous online feedback. Extensive experiments on six benchmarks demonstrate that MARS outperforms state-of-the-art self-evolving systems while significantly reducing computational overhead. Code are available at \url{https://anonymous.4open.science/r/MARS-9F16}
\end{abstract}

\section{Introduction}

Large language models (LLMs) have enabled autonomous agents capable of complex reasoning, planning, and tool use \cite{brown2020language, wei2022chain, yao2023react}. However, current agents rely on fixed, human-designed components---manually crafted prompts, predefined workflows, and static configurations---limiting their adaptability to strategies within human intuition \cite{hu2025adas, wang2024survey}. While machine learning history shows hand-designed solutions are consistently replaced by learned ones \cite{elsken2019neural, zoph2017neural}, agent development remains largely manual. The theoretical basis for self-improving AI dates back to Schmidhuber's G\"{o}del machines \cite{schmidhuber2007godel}, which formalized self-referential systems that rewrite their own code. Although formal proof requirements make the original framework impractical \cite{steunebrink2011family}, it has inspired modern self-improving systems that rely on empirical validation instead.
\begin{figure}
    \centering
    \includegraphics[width=\linewidth]{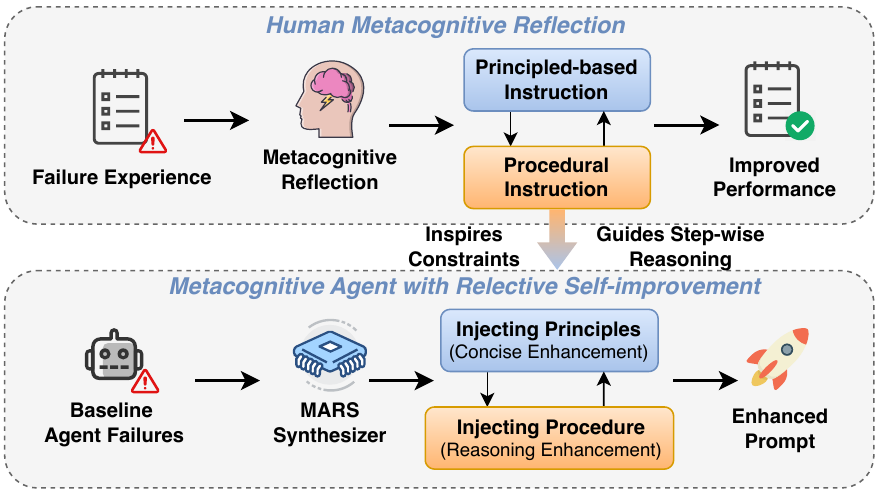}
    \caption{The Cognitive Inspiration behind MARS. This framework parallels human reflection with the MARS (Metacognitive Agent with Reflective Self-improvement) agent, converting baseline agent failures into principled-based and procedural instructions to synthesize enhanced prompts.}
    \label{fig:placeholder}
\end{figure}

However, current self-improvement frameworks for LLM agents tend to be constrained by multi-turn recursiveness, which results in inefficient learning and adaptation, as well as excessive computational resource usage. Humans, by contrast, are able to resolve previous errors and adapt to new solutions more efficiently through structured learning approaches. Research in education science has identified two complementary paradigms for guiding learners \cite{hiebert1986conceptual, anderson1983architecture}. The first is \textit{principle-based learning}, which focuses on helping learners avoid mistakes by establishing conceptual categories of what is correct versus incorrect, and understanding the underlying rules that govern a domain \cite{hiebert1986conceptual, rittle2001developing}. The second is \textit{procedural learning}, which emphasizes using prior experience and step-by-step reasoning to increase the likelihood of successful outcomes \cite{anderson1983architecture, kolb1984experiential}. Rather than learning in isolation, humans benefit most when they integrate both approaches through systematic reflection and summarization of their experiences. Studies in metacognition have shown that structured reflection---where learners explicitly analyze what worked, what failed, and why---significantly improves learning efficiency and knowledge transfer \cite{flavell1979metacognition, kaplan2013reflection, stanton2021fostering}. Furthermore, research on productive failure demonstrates that learning from one's own errors, when properly guided, leads to deeper conceptual understanding than direct instruction alone \cite{kapur2014productive, kapur2010productive}.

In this work, we propose \textbf{MARS} (\textbf{M}etacognitive \textbf{A}gent with \textbf{R}eflective \textbf{S}elf-improvement), a framework that enables multi-agent systems to achieve efficient self-improvement within a single recurrence cycle by integrating both principle-based and procedural learning approaches. Inspired by human metacognitive learning, MARS allows agents to systematically reflect on their experiences, extracting general principles that help avoid past mistakes while simultaneously deriving procedural knowledge that replicates successful strategies. Unlike existing self-evolving agent frameworks that rely on multi-turn recursive improvement, which often leads to inefficient learning and excessive computational costs, MARS consolidates the learning process through structured summarization, enabling agents to maximize adaptation efficiency in each improvement cycle.

Our main contributions are as follows:
\begin{itemize}
    \item We propose MARS, a self-improvement framework for multi-agent systems that integrates principle-based and procedural learning inspired by human meta-cognitive theory.
    \item We introduce a triple-pathway reflection mechanism that extracts: (1) normative principles for error avoidance, (2) procedural strategies for success replication, and (3) a unified synthesis of both pathways.
    \item We design a structured summarization module that consolidates learning within a single cycle, reducing computational overhead from multi-turn recursive improvement.
    \item We conduct extensive experiments on challenging knowledge and reasoning benchmarks, showing MARS outperforms existing self-evolving frameworks while requiring fewer iterations.
\end{itemize}

\section{Related Work}

Recent research has transitioned from static prompting to \textit{self-evolving agents}—systems capable of analyzing their own performance, learning from errors, and modifying their behavior to improve over time. Drawing from meta-learning principles \cite{finn2017model, hospedales2022meta}, these approaches can be broadly categorized into two paradigms: verbal reflection and structural self-modification.

The first category utilizes {verbal reflection} to facilitate learning from failure. \textit{Reflexion} \cite{shinn2023reflexion} introduced this paradigm by enabling agents to generate natural language critiques of their mistakes, storing them in episodic memory to guide future reasoning. \textit{RISE} \cite{qu2024rise} extends this by training models to iteratively detect and correct errors across multiple turns, demonstrating that self-correction capabilities can be internalized through fine-tuning. While effective, these methods primarily rely on inference-time recursion or memory retrieval rather than permanent parameter or prompt optimization.
The second category focuses on {automated architecture and code evolution}. Systems like \textit{ADAS} \cite{hu2025adas} employ meta-agents to iteratively generate and evaluate new agent designs in code, while \textit{AgentSquare} \cite{shang2025agentsquare} adopts a modular approach to evolve components for planning, reasoning, and tool use. Similarly, \textit{Agent-Pro} \cite{zhang2024agentpro} optimizes policies through reflection on historical trajectories. Taking this further, fully self-referential approaches allow agents to modify their own underlying source code. The \textit{G\"{o}del Agent} \cite{yin2025godelagent} enables agents to rewrite their logic guided by high-level objectives, a concept extended by the \textit{Darwin G\"{o}del Machine} \cite{zhang2025darwin} and \textit{SICA} \cite{robeyns2025sica}, which integrate evolutionary search to explore diverse self-improvement paths.

Despite these advancements, current frameworks face significant efficiency bottlenecks. Reflection-based methods often depend on computationally expensive multi-turn recursive loops, while code-modifying agents require complex validation environments. Unlike these approaches, our framework draws inspiration from human metacognitive theory to achieve efficient self-improvement within a \textit{single recurrence cycle}. % By integrating principle-based and procedural learning, MARS consolidates insights into optimized prompt instructions, reducing the overhead associated with continuous online feedback or evolutionary search.

\section{Methodology}

\begin{figure*}[th]
    \centering
    \includegraphics[width=0.9\linewidth]{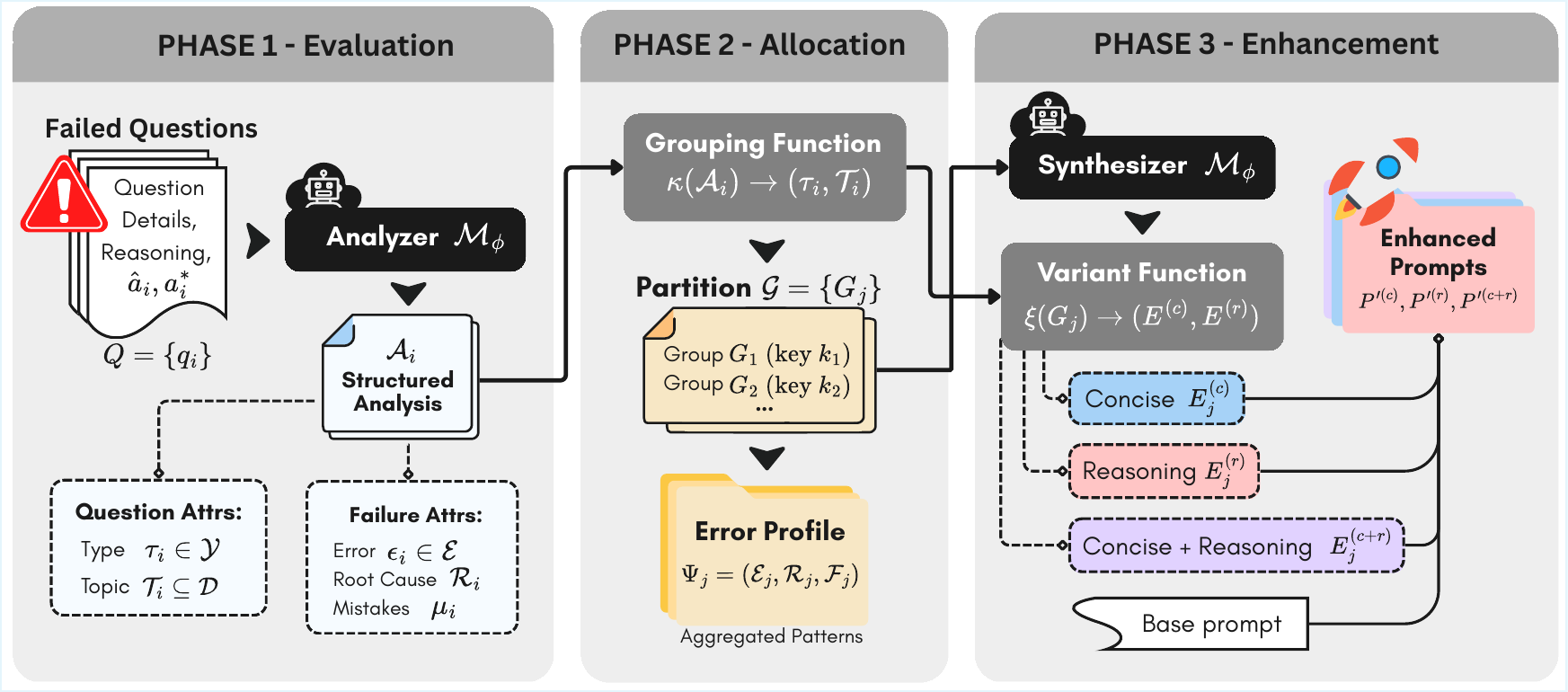}
    \caption{Overview of the proposed framework: (1) diagnose failed questions into structured analyses $\mathcal{A}_i$, (2) group by type-topic keys and aggregate error profiles $\Psi_j$, and (3) synthesize enhancements via weighted aggregation to produce $P'^{(c)}$ and $P'^{(r)}$.}
    \label{methodology}
\end{figure*}
To achieve efficient self-improvement without the computational overhead of recursive loops, we propose MARS, a three-phase framework designed to systematically transform sporadic model failures into targeted, actionable prompt enhancements. Rather than treating errors as isolated incidents, our approach aggregates failures to identify systematic weaknesses, synthesizes remediation strategies, and integrates them into a self-improving loop. Figure~\ref{methodology} illustrates the complete pipeline.

The framework operates as follows. In the \textit{Evaluation} phase, an analyzer model $\mathcal{M}_\phi$ examines each failed question and produces a structured analysis $\mathcal{A}_i$ capturing both question characteristics (type $\tau_i$, topics $\mathcal{T}_i$) and failure attributes (error type $\epsilon_i$, root cause $\rho_i$, specific mistake $\mu_i$). The \textit{Failure Allocation} phase then applies a grouping function $\kappa$ to partition analyses into groups $\mathcal{G} = \{G_j\}$ based on shared type-topic keys, aggregating diagnostic attributes into group-level error profiles $\Psi_j$. Finally, the \textit{Enhancement Generation} phase synthesizes targeted enhancements $(E_j^{(c)}, E_j^{(r)})$ for each group and combines them with the base prompt $P$ via weighted aggregation to produce enhanced prompts $P'^{(c)}$ and $P'^{(r)}$. We detail each phase below.

\subsection{Evaluation}

The first phase of our enhancement pipeline performs fine-grained diagnosis of each incorrectly answered question. Rather than treating failures as a homogeneous set, we analyze each instance independently to understand the precise reasoning breakdown that led to the incorrect response.

Formally, let $\mathcal{Q} = \{q_i\}_{i=1}^{n}$ denote a set of failed questions from the benchmark evaluation. Each instance $q_i$ comprises question text, options, ground-truth answer $a_i^*$, the model's predicted answer $\hat{a}_i$, and generated reasoning trace. We employ a specialized analyzer model, $\mathcal{M}_\phi$, to dissect these components. For every $q_i$, the analyzer produces a structured analysis 
\begin{equation}
\mathcal{A}_i = (\tau_i, \mathcal{T}_i, \epsilon_i, \rho_i, \mu_i)
\label{eq:analysis}
\end{equation}
, which encapsulates two distinct categories of attributes: % We employ a separate analyzer model $\mathcal{M}_\phi$ to examine each failure. For every failed question, the analyzer receives the complete context and produces a structured analysis $\mathcal{A}_i = (\tau_i, \mathcal{T}_i, \epsilon_i, \rho_i, \mu_i)$ containing two categories of attributes.

The first category characterizes question itself. It assigns a ``question type'' $\tau_i \in \mathcal{Y}$ (where $\mathcal{Y} = \{\text{factual, conceptual, calculation, application}\}$) and identifies a set of ``topics'' $\mathcal{T}_i \subseteq \mathcal{D}$ derived from domain vocabulary $\mathcal{D}$. As detailed in Table~\ref{tab:question_taxonomy}, the combination of question type $\tau_i$ and topic $\mathcal{T}_i$ serves as composite key for grouping failures in the subsequent Allocation phase.
% The first category characterizes the question itself: the ``question type'' $\tau_i \in \mathcal{Y}$ where $\mathcal{Y} = \{\text{factual, conceptual, calculation, application}\}$, and a set of ``topics'' $\mathcal{T}_i \subseteq \mathcal{D}$ representing the specific subject matter drawn from the domain vocabulary $\mathcal{D}$.
The second category characterizes specific error mechanism. This includes an ``error type'' $\epsilon_i \in \mathcal{E}$, a natural language ``root cause'' $\rho_i$ explaining the fundamental reasoning deficit, and a ``specific mistake'' $\mu_i$ pinpointing the exact step where logic diverged. The error taxonomy $\mathcal{E}$, presented in Table~\ref{tab:error_taxonomy}, defines six categories ranging from conceptual misunderstandings to calculation errors.
% The second category characterizes the failure: an ``error type'' $\epsilon_i \in \mathcal{E}$ drawn from a predefined taxonomy, a natural language ``root cause'' $\rho_i$ explaining the fundamental reasoning failure, and the ``specific mistake'' $\mu_i$ identifying the exact step where reasoning diverged.

% Table~\ref{tab:question_taxonomy} describes the question type categories used to characterize each failed question. Together with the topic set $\mathcal{T}_i$, the question type $\tau_i$ forms the composite key for failure allocation.

\begin{table*}[t]
\centering
\resizebox{\linewidth}{!}{
\begin{tabular}{lll}
\toprule
\textbf{Type} & \textbf{Description} & \textbf{Example} \\
\midrule
Factual & Recall of specific facts or definitions & ``What is the atomic number of carbon?'' \\
Conceptual & Understanding of principles or theories & ``Why does entropy increase in isolated systems?'' \\
Calculation & Quantitative problem requiring computation & ``Calculate the final velocity given...'' \\
Application & Applying knowledge to novel scenarios & ``Which treatment would be most effective for...'' \\
Analysis & Breaking down complex information & ``What can be inferred from this experimental data?'' \\
Comparison & Evaluating similarities or differences & ``Which compound has higher boiling point?'' \\
\bottomrule
\end{tabular}%
}
\caption{Question type categories.}
\label{tab:question_taxonomy}
\end{table*}

To ensure consistent classification, we enforce a strict exclusivity rule: each failure is assigned to exactly one category in $\mathcal{E}$. In cases where multiple failure modes co-occur (e.g., a calculation error stemming from a conceptual misunderstanding), the analyzer assigns the category corresponding to the \textit{earliest point of divergence} in the reasoning chain. The final output of this phase is the collection of structured analyses $\mathbb{A} = \{\mathcal{A}_i\}_{i=1}^{n}$.
% The error taxonomy $\mathcal{E}$ defines six mutually exclusive categories: conceptual misunderstanding, calculation error, misreading of question or choices, incomplete analysis, incorrect elimination of candidates, and knowledge gap, as presented in Table~\ref{tab:error_taxonomy}. This controlled vocabulary ensures consistent classification and enables systematic pattern discovery in subsequent phases. The output of this phase is a collection of structured analyses $\mathbb{A} = \{\mathcal{A}_i\}_{i=1}^{n}$ that serve as input to the allocation step.

\begin{table*}[t]
\centering
\resizebox{\linewidth}{!}{
\begin{tabular}{lll}
\toprule
\textbf{Category} & \textbf{Description} & \textbf{Typical Manifestation} \\
\midrule
Conceptual Misunderstanding & Fundamental confusion about domain principles & Misapplying laws; confusing related concepts \\
Calculation Error & Computational or mathematical mistakes & Arithmetic errors; unit conversion mistakes \\
Misreading & Misinterpretation of question or choices & Overlooking negation; misidentifying the question \\
Incomplete Analysis & Premature termination of reasoning & Stopping after partial solution \\
Wrong Elimination & Incorrect rejection of candidate answers & Eliminating correct answer based on flawed logic \\
Knowledge Gap & Absence of requisite domain knowledge & Missing facts; unfamiliar terminology \\
\bottomrule
\end{tabular}%
}
\caption{Error categories and descriptions.}
\label{tab:error_taxonomy}
\end{table*}

% These categories are mutually exclusive; each failure is assigned to exactly one category based on the primary cause identified by the analyzer model $\mathcal{M}_\phi$. When multiple failure modes co-occur, the analyzer assigns the category corresponding to the earliest point of divergence in the reasoning chain.

\subsection{Failure Allocation}

This aggregation transforms sparse, per-instance observations into dense, group-level patterns. By clustering failures with shared characteristics, the allocation phase enables the subsequent system to generate high-level guidance that addresses classes of errors simultaneously, ensuring that the enhanced prompts are both targeted and scalable.

The second phase organizes individual error analyses into semantically coherent groups to enable pattern discovery. While the previous phase examined each failure in isolation, this phase identifies structural similarities across failures that may share common remediation strategies. We define a composite grouping function 
\begin{equation}
\kappa: \mathbb{A} \rightarrow \mathcal{Y} \times 2^{\mathcal{D}}, \quad \kappa(\mathcal{A}_i) = (\tau_i, \mathcal{T}_i)
\label{eq:grouping}
\end{equation} 
that maps each analysis to its ``type-topic key'' via $\kappa(\mathcal{A}_i) = (\tau_i, \mathcal{T}_i)$. This two-dimensional grouping captures the intuition that errors on calculation questions about thermodynamics likely stem from different causes than errors on conceptual questions about molecular biology, and thus require distinct enhancement strategies.

Given the analysis set $\mathbb{A}$ from the previous phase, we construct a partition \begin{equation}
\Psi_j = (\mathcal{E}_j, \mathcal{R}_j, \mathcal{F}_j)
\label{eq:error_profile}
\end{equation}
, where each group $G_j = \{\mathcal{A}_i \in \mathbb{A} : \kappa(\mathcal{A}_i) = k_j\}$ contains all analyses sharing the same type-topic key $k_j$. Within each group, we aggregate the diagnostic attributes to form a collective ``error profile'' $\Psi_j = (\mathcal{E}_j, \mathcal{R}_j, \mathcal{F}_j)$, comprising the set of observed error types $\mathcal{E}_j = \{\epsilon_i : \mathcal{A}_i \in G_j\}$, recurring root causes $\mathcal{R}_j = \{\rho_i : \mathcal{A}_i \in G_j\}$, and common difficulty factors $\mathcal{F}_j$. 
% This aggregation transforms sparse per-instance observations into denser group-level patterns that characterize systematic failure modes.
% The allocation phase serves as a bridge between individual diagnosis and enhancement generation. By clustering failures with shared characteristics, we enable the subsequent phase to synthesize targeted guidance that addresses multiple related errors simultaneously rather than treating each failure independently.

This aggregation transforms sparse, per-instance observations into dense, group-level patterns. By clustering failures with shared characteristics, the allocation phase enables the subsequent system to generate high-level guidance that addresses classes of errors simultaneously, ensuring that the enhanced prompts are both targeted and scalable.

\begin{table*}[th]
\centering
\resizebox{\linewidth}{!}{
\begin{tabular}{ll llll}
\toprule
\multirow{2}{*}{\textbf{Benchmark}} & \multirow{2}{*}{\textbf{Focus}} & \multicolumn{2}{c}{\textbf{Categories}} & \multicolumn{2}{c}{\textbf{Details}} \\
\cmidrule(lr){3-4} \cmidrule(lr){5-6}
 & & Primary Domains & Sub-categories & Key Features & Size \\
\midrule
DROP \cite{dua2019drop} & Discrete Reasoning & Numerical Operations & Add/Sub, Min/Max, Count, Select, Compare & NFL \& History passages & 96K \\
\midrule
MGSM \cite{shi2022language} & Multilingual Math & Grade-school Math & Arithmetic, Word Problems, Algebra & 11 languages (BN, SW, TE incl.) & 250 \\
\midrule
 & & STEM & Physics, Chemistry, CS, Math, Biology & & \\
 & & Humanities & History, Philosophy, Law, Ethics & & \\
\multirow{-3}{*}{MMLU \cite{hendrycks2021measuring}} & \multirow{-3}{*}{General Knowledge} & Social Sciences & Economics, Psychology, Politics & \multirow{-3}{*}{57 subjects; Elem.--Prof.} & \multirow{-3}{*}{15.9K} \\
\midrule
 & & Biology & Molecular Bio (8\%), Genetics & & \\
 & & Physics & General (10\%), Electromagnetism & & \\
\multirow{-3}{*}{GPQA \cite{rein2024gpqa}} & \multirow{-3}{*}{Graduate Science} & Chemistry & Organic (36\%), General & \multirow{-3}{*}{PhD-level; Google-proof} & \multirow{-3}{*}{448} \\
\midrule
 & & Mathematics (41\%) & Algebra, Analysis, Combinatorics & & \\
 & & Sciences (27\%) & Physics, Chemistry, Biology & & \\
\multirow{-3}{*}{HLE \cite{phan2025humanitysexam}} & \multirow{-3}{*}{Expert Academic} & Tech \& Humanities (32\%) & CS/AI, Engineering, Social Sci. & \multirow{-3}{*}{100+ areas; 14\% multimodal} & \multirow{-3}{*}{2.5K} \\
\midrule
 & & Algebra \& Number Theory & Linear, Abstract, Primes, Modular & & \\
 & & Geometry & Euclidean, Analytic, Projective & & \\
\multirow{-3}{*}{Omni-MATH \cite{gao2024omnimath}} & \multirow{-3}{*}{Olympiad Math} & Analysis \& Discrete & Calculus, Combinatorics, Graph Theory & \multirow{-3}{*}{33+ sub-domains; 10 levels} & \multirow{-3}{*}{4.4K} \\
\bottomrule
\end{tabular}}
\caption{Question categories across six LLM evaluation benchmarks used for category-based hybrid enhancement.}
\label{tab:benchmark_categories}
\end{table*}

\subsection{Enhancement Generation}

The final phase synthesizes the group-level error patterns into prompt enhancements that guide the model toward correct reasoning. For each group $G_j \in \mathcal{G}$, we first perform pattern analysis to extract actionable guidance, then integrate this guidance into the original prompt template.

Given a group $G_j$ with its aggregated error profile $\Psi_j$, we query the analyzer model $\mathcal{M}_\phi$ to synthesize targeted remediation strategies. The analyzer examines the common error types $\mathcal{E}_j$, shared root causes $\mathcal{R}_j$, and recurring mistakes within the group, then produces structured guidance including typical pitfalls to avoid, verification steps to perform, and domain-specific reasoning strategies. This group-level synthesis captures patterns that may not be apparent from any single failure but emerge clearly when examining multiple related errors.

We define an enhancement generation function \begin{equation}
\xi: \mathcal{G} \rightarrow \mathcal{S}^{(c)} \times \mathcal{S}^{(r)}, \quad \xi(G_j) = (E_j^{(c)}, E_j^{(r)})
\label{eq:enhancement_function}
\end{equation}
that produces two distinct enhancement variants to accommodate different reasoning scenarios. The ``concise'' enhancement $E_j^{(c)} \in \mathcal{S}^{(c)}$ provides brief warnings and key points, suitable for quick reference during inference. The ``reasoning'' enhancement $E_j^{(r)} \in \mathcal{S}^{(r)}$ supplies minimal hints designed to trigger self-correction without over-constraining the reasoning process. Formally, for each group $G_j$, we produce an enhancement pair $\xi(G_j) = (E_j^{(c)}, E_j^{(r)})$.

The final enhanced prompt $P'$ is constructed by appending the relevant enhancements to the base prompt $P$. We define an aggregation operator $\bigoplus$ that combines enhancements weighted by group cardinality $|G_j|$, prioritizing guidance derived from larger groups where more failures share the same type-topic characteristics. The resulting enhanced prompts are given by {\small
\begin{equation}
P'^{(c)} = P \oplus \bigoplus_{j=1}^{m} w_j E_j^{(c)}, \quad P'^{(r)} = P \oplus \bigoplus_{j=1}^{m} w_j E_j^{(r)}
\label{eq:final_prompt}
\end{equation}
}
, where $w_j \propto |G_j|$, each embedding the collective remediation knowledge extracted from the failure analysis pipeline.

Beyond applying enhancements uniformly, we introduce a \textbf{Hybrid} strategy that dynamically selects the optimal enhancement type per question category. We apply four strategies: Concise, Reasoning, Concise+Reasoning, and Hybrid. The first three serve as ablation baselines. For the Hybrid strategy, each dataset is partitioned into train, validation, and test splits (8:1:1). The training set generates enhancements via Phases 1-3. The validation set determines which enhancement type performs best for each category $c$:
\begin{equation}
E^*_c = \operatorname*{arg\,max}_{E \in \{E^{(c)}, E^{(r)}, E^{(c+r)}\}} \text{Acc}(E, \mathcal{V}_c)
\label{eq:hybrid_selection}
\end{equation}
where $\mathcal{V}_c$ denotes validation questions in category $c$ and $E^{(c+r)}$ combines both enhancement types. The selected $E^*_c$ is applied to matching test questions.

\begin{table*}[th]
\centering
\resizebox{0.8\linewidth}{!}{
\begin{tabular}{ll llll}
\toprule
\multirow{2}{*}{\textbf{Method}} & \multirow{2}{*}{\textbf{Enhancement}} & \multicolumn{2}{c}{\textbf{Reasoning}} & \multicolumn{2}{c}{\textbf{Knowledge}} \\
\cmidrule(lr){3-4} \cmidrule(lr){5-6}
 & & DROP & MGSM & MMLU & GPQA \\
\midrule
MetaAgentSearch \cite{hu2025automateddesignagenticsystems} & - & 79.4$^\dagger$ & 53.4$^\dagger$ & 69.6$^\dagger$ & 34.6$^\dagger$ \\
Gödel Agent \cite{yin2025godelagent} & - & 80.9$^\dagger$ & 64.2$^\dagger$ & 70.9$^\dagger$ & 34.9$^\dagger$ \\
\midrule 
\multirow{5}{*}{Zero-shot \cite{brown2020language}} 
 & n/a & 62.0 & 35.0 & 64.0 & 11.8 \\
 & Concise & 63.5\smark{+1.5} & 37.6\smark{+2.6} & 60.7\smark{-3.3} & 11.8\smark{+0.0} \\
 & Reasoning & 65.2\smark{+3.2} & 37.0\smark{+2.0} & 64.0\smark{+0.0} & 12.7\smark{+0.9} \\
 & Concise+Reasoning & 63.8\smark{+1.8} & 35.2\smark{+0.2} & 64.2\smark{+0.2} & 12.7\smark{+0.9} \\
 \rowcolor{gray!20}& Hybrid & 68.4\smark{+6.4} & 39.4\smark{+4.4} & 65.1\smark{+1.1} & 20.0\smark{+8.2} \\
\midrule
\multirow{5}{*}{Zero-shot-CoT \cite{kojima2022large}} 
 & n/a & 74.5 & 52.9 & 65.8 & 16.4 \\
 & Concise & 77.2\smark{+2.7} & 54.4\smark{+1.5} & 66.8\smark{+1.0} & 19.1\smark{+2.7} \\
 & Reasoning & 78.8\smark{+4.3} & 54.1\smark{+1.2} & 69.0\smark{+3.2} & 19.1\smark{+2.7} \\
 & Concise+Reasoning & 78.1\smark{+3.6} & 54.6\smark{+1.7} & 69.0\smark{+3.2} & 17.3\smark{+0.9} \\
 \rowcolor{gray!20}& Hybrid & \textbf{81.6}\smark{+7.1} & 56.4\smark{+3.5} & 70.5\smark{+4.7} & 22.2\smark{+5.8} \\
\midrule
\multirow{5}{*}{Self-Refine \cite{madaan2023selfrefine}} 
 & n/a & 77.8 & 57.5 & 48.8 & \textbf{36.4}\\
 & Concise & 80.5\smark{+2.7} & 60.5\smark{+3.0} & 60.5\smark{+11.7} & \textbf{38.2}\smark{+1.8} \\
 & Reasoning & \textbf{82.1}\smark{+4.3} & 58.4\smark{+0.9} & 59.0\smark{+10.2} & \textbf{40.9}\smark{+4.5} \\
 & Concise+Reasoning & \textbf{81.2}\smark{+3.4} & 58.7\smark{+1.2} & 63.9\smark{+15.1} & 32.7\smark{-3.7} \\
 \rowcolor{gray!20}& Hybrid & \textbf{84.3}\smark{+6.5} & 61.3\smark{+3.8} & 64.6\smark{+15.8} & \textbf{49.1}\smark{+12.7} \\
\midrule
\multirow{5}{*}{Self-Consistency \cite{wang2023selfconsistency}} 
 & n/a & 79.5 & 63.5 & 61.8 & 18.2 \\
 & Concise & \textbf{83.8}\smark{+4.3} & \textbf{73.7}\smark{+10.2} & 69.3\smark{+7.5} & 26.4\smark{+8.2} \\
 & Reasoning & \textbf{84.5}\smark{+5.0} & \textbf{73.2}\smark{+9.7} & \textbf{71.3}\smark{+9.5} & 24.6\smark{+6.4} \\
 & Concise+Reasoning & \textbf{83.2}\smark{+3.7} & \textbf{73.7}\smark{+10.2} & \textbf{71.7}\smark{+9.9} & 21.8\smark{+3.6} \\
 \rowcolor{gray!20}& Hybrid & \textbf{86.2}\smark{+6.7} & \textbf{74.3}\smark{+10.8} & \textbf{72.5}\smark{+10.7} & 33.6\smark{+15.4} \\
\bottomrule
\end{tabular}}
\caption{Results on DROP, MGSM, MMLU, and GPQA benchmarks. DROP uses F1 score; others use accuracy (\%). Subscripts indicate improvement over n/a baseline. \textbf{Bold} indicates results exceeding Gödel Agent. $^\dagger$ are results obtained from \citealp{yin2025godelagent}.}
\label{tab:main_results}
\end{table*}
\section{Experiments}
\paragraph{Datasets.}
We evaluate on six benchmarks spanning reasoning capacity and knowledge coverage. For reasoning capacity, we use DROP~\cite{dua2019drop}, a reading comprehension benchmark requiring discrete reasoning operations such as addition, counting, and sorting; MGSM~\cite{shi2022language}, the Multilingual Grade School Math benchmark containing 250 problems; and OMNI-math~\cite{gao2024omnimath}, an Olympiad-level benchmark with 4,428 competition problems across 33 sub-domains. For knowledge coverage, we use MMLU~\cite{hendrycks2021measuring}, which spans 57 subjects including STEM, humanities, and social sciences; GPQA~\cite{rein2024gpqa}, a graduate-level ``Google-proof'' Q\&A benchmark with expert-written questions in biology, physics, and chemistry; and HLE~\cite{phan2025humanitysexam} (Humanity's Last Exam), a frontier benchmark with 2,500 expert-level questions designed to test the limits of current AI systems.

For DROP, MGSM, MMLU, and GPQA, we use \texttt{gpt-3.5-turbo} for comparison with baselines. For the more challenging benchmarks, Omni-MATH and Humanity's Last Exam, we use \texttt{gpt-4o} due to its stronger reasoning capabilities and more recent training data cutoff.

\paragraph{Implementation.}
We evaluate four base prompting methods: Zero-shot~\cite{brown2020language}, Zero-shot-CoT~\cite{kojima2022large} which appends ``Let's think step by step'' to elicit reasoning, Self-Refine~\cite{madaan2023selfrefine} which iteratively critiques and improves responses, and Self-Consistency~\cite{wang2023selfconsistency} which samples multiple reasoning paths and selects answers via majority voting.

We apply four enhancement strategies to each prompting method: Concise (principle-based do's and don'ts for avoiding common errors), Reasoning (explicit step-by-step instructions to follow correct rationale), Concise+Reasoning (combining both), and Hybrid (dynamically selecting strategies based on question category). Concise, Reasoning, and Concise+Reasoning serve as ablation baselines to isolate the contribution of each enhancement type. For the Hybrid strategy, we leverage the question categories outlined in Table~\ref{tab:benchmark_categories}: DROP (5 discrete reasoning types), MGSM (3 math types × 11 languages), MMLU (57 subjects across 4 domains), GPQA (3 scientific domains), Humanity's Last Exam (8 broad categories), and Omni-MATH (33+ mathematical sub-domains). Each dataset is split into train:val:test = 8:1:1. The validation set is used to discover the optimal enhancement strategy for each question category, which is then applied to the corresponding categories in the test set. We compare against MategAgentSearch and G\"{o}del Agent~\cite{yin2025godelagent}, in which G\"{o}del Agent represents a state-of-the-art meta-learning optimized agent system. All experiments use temperature $T=0$ and maximum token length of 3,000. For Self-Consistency, we sample $n=5$ responses with temperature $T=0.7$. We report F1 score for DROP and accuracy for all other benchmarks.

\begin{figure*}[th]
\centering
\begin{subfigure}[b]{0.48\textwidth}
    \centering
    \includegraphics[height=5cm]{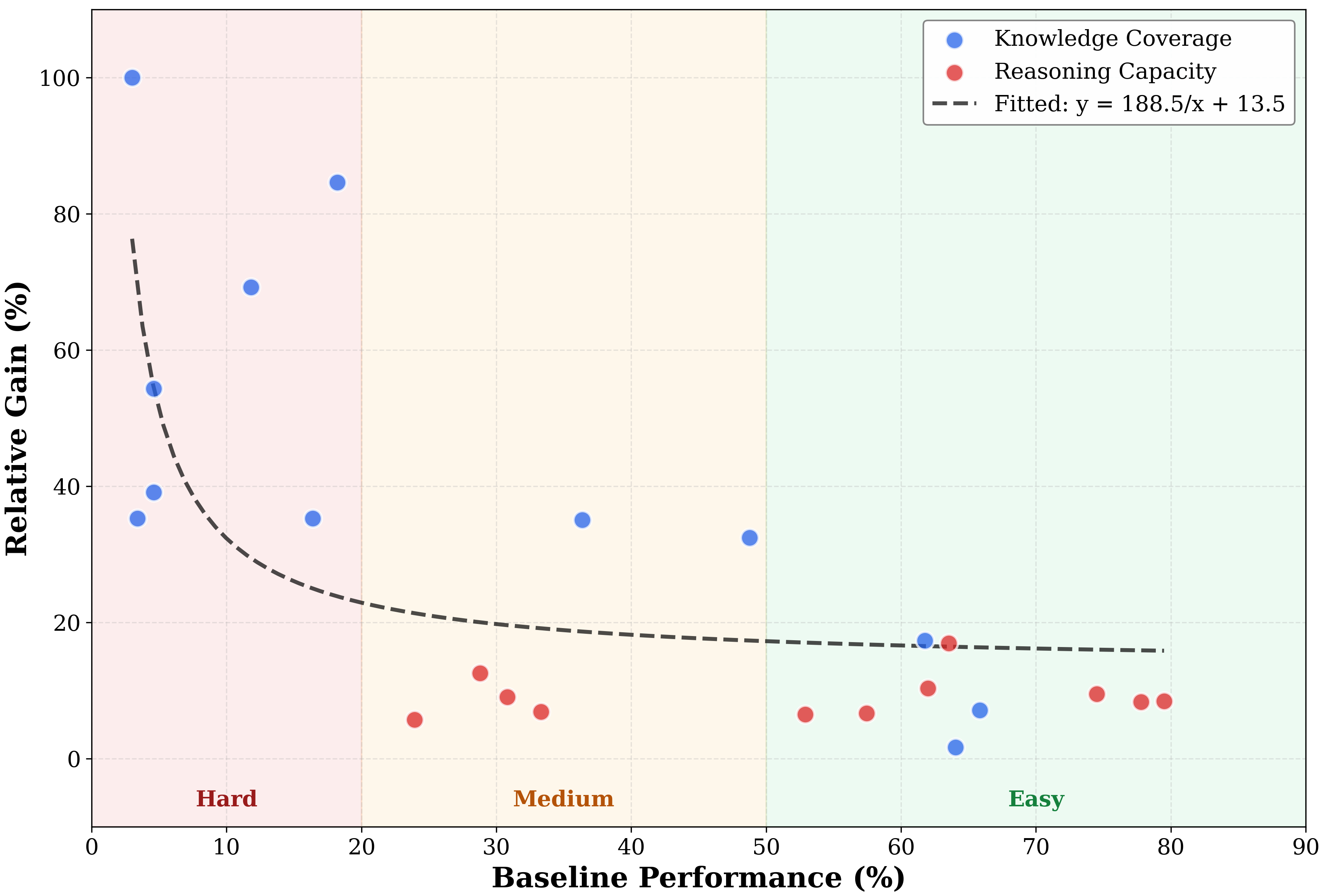}
    \caption{Baseline performance vs. relative gain from hybrid enhancement. Background shading indicates difficulty tiers.}
    \label{fig:gain}
\end{subfigure}
\hfill
\begin{subfigure}[b]{0.48\textwidth}
    \centering
    \includegraphics[height=5.3cm]{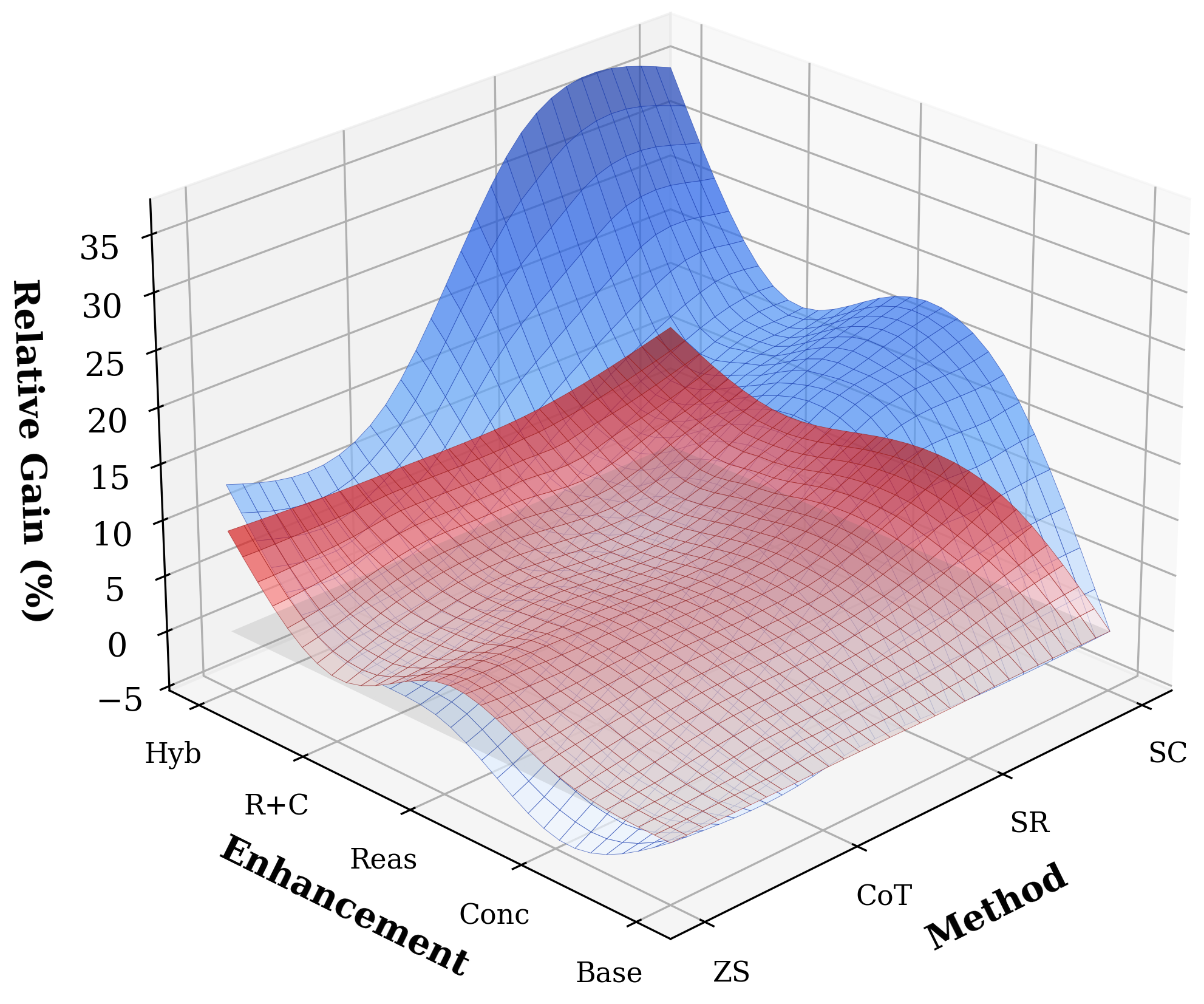}
    \caption{3D surface of relative performance gain across prompting methods (X), enhancement types (Y), and accuracy (Z).}
    \label{fig:category}
\end{subfigure}
\caption{Performance gain analysis. (a) Inverse relationship between task difficulty and enhancement effectiveness. (b) Performance landscape comparison between Knowledge Coverage (blue) and Reasoning Capacity (red).}
\label{fig:enhancement_analysis}
\end{figure*}

\begin{table}[t]
\centering
\resizebox{\linewidth}{!}{
\begin{tabular}{ll ll}
\toprule
\multirow{2}{*}{\textbf{Method}} & \multirow{2}{*}{\textbf{Enhancement}} & \textbf{Reasoning} & \textbf{Knowledge} \\
\cmidrule(lr){3-3} \cmidrule(lr){4-4}
 & & OMNI-math & HLE \\
\midrule
\multirow{5}{*}{Zero-shot} 
 & n/a & 23.93 & 3.40 \\
 & Concise & 23.44\smark{-0.49} & 3.20\smark{-0.20} \\
 & Reasoning & 24.56\smark{+0.63} & 3.40\smark{+0.00} \\
 & R+C $\star$ & 22.43\smark{-1.50} & 4.20\smark{+0.80} \\
 \rowcolor{gray!20}& Hybrid & 25.30\smark{+1.37} & 4.60\smark{+1.20} \\
\midrule
\multirow{5}{*}{Zero-shot-CoT} 
 & n/a & 30.81 & 4.60 \\
 & Concise & 31.04\smark{+0.23} & 5.10\smark{+0.50} \\
 & Reasoning & 31.83\smark{+1.02} & 5.40\smark{+0.80} \\
 & R+C $\star$ & 31.72\smark{+0.91} & 5.20\smark{+0.60} \\
 \rowcolor{gray!20}& Hybrid & 33.60\smark{+2.79} & 6.40\smark{+1.80} \\
\midrule
\multirow{5}{*}{Self-Refine} 
 & n/a & 28.78 & 4.60 \\
 & Concise & 28.67\smark{-0.11} & 6.40\smark{+1.80} \\
 & Reasoning & 30.59\smark{+1.81} & 5.50\smark{+0.90} \\
 & R+C $\star$ & 29.12\smark{+0.34} & 6.00\smark{+1.40} \\
 \rowcolor{gray!20}& Hybrid & 32.40\smark{+3.62} & \textbf{7.10}\smark{+2.50} \\
\midrule
\multirow{5}{*}{Self-Consistency} 
 & n/a & 33.30 & 3.00 \\
 & Concise & 33.60\smark{+0.30} & 3.40\smark{+0.40} \\
 & Reasoning & 34.60\smark{+1.30} & 5.20\smark{+2.20} \\
 & R+C $\star$ & 32.50\smark{-0.80} & 4.80\smark{+1.80} \\
 \rowcolor{gray!20}& Hybrid & \textbf{35.60}\smark{+2.30} & 6.00\smark{+3.00} \\
\bottomrule
\end{tabular}}
\caption{Results on OMNI-math and HLE. Accuracy (\%) reported. Subscripts indicate improvement over baseline. \textbf{Bold} indicates best result per dataset. $\star$s are short for reasoning plus concise enhancements}
\label{tab:additional_results}
\end{table}
\section{Results and Analysis}

\paragraph{Main Results}

Table~\ref{tab:main_results} presents results across four benchmarks: reasoning capacity (DROP and MGSM) and knowledge coverage (MMLU and GPQA). We compare our enhancement strategies against MetaAgentSearch~\cite{hu2025automateddesignagenticsystems} and Gödel Agent~\cite{yin2025godelagent}. MetaAgentSearch and Gödel Agent employ instruction-free self-improvement through multiple recursive iterations, automatically discovering agent architectures without human-crafted guidance. In contrast, methods like Self-Refine~\cite{madaan2023selfrefine} rely on human-crafted prompts encoding explicit refinement strategies without groundtruth. Notably, even without our enhancements, Self-Refine (36.4\%) already outperforms both Gödel Agent (34.9\%) and MetaAgentSearch (34.6\%) on GPQA. This demonstrates that well-designed human-crafted prompting can surpass instruction-free recursive optimization. With hybrid enhancement, Self-Consistency surpasses Gödel Agent on three benchmarks (DROP: 86.2 vs.\ 80.9, MGSM: 74.3 vs.\ 64.2, MMLU: 72.5 vs.\ 70.9), while Self-Refine with hybrid reaches 49.1\% on GPQA. These results suggest that MARS offers a cost-effective alternative to complex recursive agent systems.

Zero-shot and Zero-shot-CoT show consistent improvements, with hybrid yielding +6.4 and +7.1 on DROP respectively. Self-Refine achieves dramatic gains on knowledge benchmarks: +15.8 on MMLU and +12.7 on GPQA. Self-Consistency benefits substantially across all benchmarks (+10.8 on MGSM, +10.7 on MMLU), indicating synergy between multiple reasoning paths and enhanced prompts. Hybrid enhancement consistently yields the largest improvements, validating category-aware selection. Reasoning generally outperforms Concise on reasoning-intensive tasks. Combining both (Concise+Reasoning) does not always yield additive benefits---on GPQA with Self-Refine, it underperforms individual enhancements (32.7\% vs.\ 40.9\% for Reasoning), suggesting interference when naively combining strategies.

\paragraph{Generalization to Challenging Benchmarks}

To investigate whether MARS generalizes to more challenging evaluation settings, we evaluate on two additional benchmarks: Omni-MATH and Humanity's Last Exam (HLE). Table~\ref{tab:additional_results} presents these results.

Omni-MATH tests advanced mathematical reasoning with baseline accuracies below 35\%. Despite this difficulty, our enhancements provide consistent improvements. Reasoning enhancement proves particularly effective, yielding gains across all prompting methods (+0.63 for Zero-shot, +1.02 for Zero-shot-CoT, +1.81 for Self-Refine, +1.30 for Self-Consistency). Hybrid enhancement achieves the best results, with Self-Consistency reaching 35.60\% (+2.30). Notably, Concise+Reasoning often underperforms individual enhancements (e.g., 32.50\% vs.\ 34.60\% for Self-Consistency), consistent with the interference pattern observed in main results.

HLE represents an extremely challenging benchmark with baseline accuracies below 5\%. Even in this difficult regime where models operate near floor-level performance, MARS demonstrates effectiveness. Self-Refine with hybrid enhancement achieves 7.10\%, a 54.3\% relative improvement over baseline (4.60\%). Zero-shot-CoT with hybrid reaches 6.40\% (+1.80), and Self-Consistency with Reasoning achieves 5.20\% (+2.20). These gains indicate that our category-aware enhancements extract meaningful improvements even on problems designed to challenge state-of-the-art systems. Finally, using a different model for enhancement generation did not result in significant changes, as presented in Appendix~\ref{appendix:qwen_results}. Qwen2.5-72B-Instruct-Turbo produced comparable enhancement patterns across both knowledge coverage and reasoning capacity benchmark that confirms MARS enhancement's effectiveness generalizes across enhancement generators rather than being dependent on a specific model.

\paragraph{Performance Gain Analysis}

We analyze the relationship between baseline performance and relative gain from prompt enhancement using scatter plots and 3D surface visualizations across knowledge coverage (HLE, GPQA, MMLU) and reasoning capacity (OMNI-math, MGSM, DROP) benchmarks (Figure \ref{fig:enhancement_analysis}).

A significant inverse correlation exists between baseline performance and relative gain (Spearman $\rho = -0.654$, $p < 0.001$), with the fitted model $\text{gain} = 188.54/\text{baseline} + 13.48$ ($R^2 = 0.443$). Critically, this relationship is category-dependent: knowledge coverage datasets exhibit strong correlation ($\rho = -0.795$, $p = 0.002$), while reasoning capacity datasets show no significant relationship ($\rho = 0.264$, $p = 0.433$). This divergence suggests fundamentally different enhancement mechanisms---knowledge tasks benefit disproportionately at low baselines, whereas reasoning gains remain uniform across difficulty levels.

The 3D surface analysis reveals a method-enhancement interaction specific to reasoning tasks. For reasoning datasets, self-consistency combined with one-turn MARS enhancement produces significantly amplified gains compared to other prompting methods (7.31\% vs.\ 2.66\%; Mann-Whitney $U$, $p = 0.050$). This amplification pattern is \textit{not observed} in knowledge coverage datasets, where high variance ($\sigma = 21.28\%$) obscures potential interaction effects and gains are primarily explained by the baseline-gain relationship. 

\begin{table}[th]
\centering
\resizebox{\linewidth}{!}{
\label{tab:gain_significance}
\begin{tabular}{lcc}
\toprule
\textbf{Items} & \textbf{Statistic} & \textbf{$p$-value} \\
\midrule
Overall baseline-gain correlation & $\rho = -0.654$ & $< 0.001$*** \\
Knowledge baseline-gain correlation & $\rho = -0.795$ & $= 0.002$** \\
Reasoning baseline-gain correlation & $\rho = 0.264$ & $= 0.433$ \\
SC amplification (Reasoning) & $U$ & $= 0.050$* \\
Categories differ in gain & $U$ & $= 0.003$** \\
\bottomrule
\end{tabular}}
\caption{Statistical significance summary for gain analysis.}
\end{table}

These findings suggest category-aware enhancement strategies: for knowledge tasks, prioritize low-baseline scenarios where gains are maximized; for reasoning tasks, leverage self-consistency methods which uniquely amplify MARS enhancement effects.

\section{Conclusion}
We presented MARS, a metacognitive framework that integrates principle-based reflection (learning what to avoid) with procedural reflection (learning how to succeed) for efficient self-improvement in LLM agents.
Existing instruction-free self-improvement methods rely on multi-turn recursive optimization that is both computationally expensive and often underperforming. MARS overcomes both shortcomings by consolidating learning into a single recurrence cycle through structured summarization, while generating targeted, category-aware enhancements from systematic failure analysis.
Experiments across six benchmarks demonstrate that MARS consistently outperforms state-of-the-art self-evolving systems with significantly reduced computational overhead, suggesting that human-inspired learning paradigms offer a practical alternative to resource-intensive recursive self-improvement.

\section*{Limitations}
Two main limitations warrant consideration. First, the predefined error taxonomy may not generalize to all task types. Our six error categories and type-topic groupings are designed for structured benchmarks with clear correct answers. For open-ended or creative tasks where errors are less categorical and more subjective, these classifications may prove insufficient. Extending MARS to such domains would require developing more flexible categorization schemes.

Second, single-cycle learning trades depth for efficiency. While our approach significantly reduces computational cost compared to recursive methods, it inherently limits the improvement achievable in one pass. For applications prioritizing maximum performance over efficiency, we recommend applying MARS iteratively—using enhanced prompts to generate new failure sets, then deriving further refinements from residual errors.
\bibliography{custom}

\appendix

\section{Algorithm}
Algorithm~\ref{alg:mars} presents the main MARS pipeline, which takes a set of failed questions and produces enhanced prompts.

\begin{algorithm}[h]
\caption{MARS Enhancement Pipeline}
\label{alg:mars}
\begin{algorithmic}[1]

\Require Failed questions $\mathcal{Q} = \{q_1, \ldots, q_n\}$, ground truths $\{a_i^*\}$, predictions $\{\hat{a}_i\}$, base prompt $P$, analyzer $\mathcal{M}_\phi$, error taxonomy $\mathcal{E}$
\Ensure Enhanced prompts $(P'^{(c)}, P'^{(s)}, P'^{(r)})$

\Statex
\State \Comment{Phase 1: Evaluation}
\State $\mathbb{A} \leftarrow \emptyset$
\For{$q_i \in \mathcal{Q}$}
    \State $\mathcal{A}_i \leftarrow \textsc{Diagnose}(\mathcal{M}_\phi, q_i, a_i^*, \hat{a}_i)$
    \State $\mathbb{A} \leftarrow \mathbb{A} \cup \{\mathcal{A}_i\}$
\EndFor

\Statex
\State \Comment{Phase 2: Failure Allocation}
\State $\mathcal{G} \leftarrow \textsc{Cluster}(\mathbb{A})$

\Statex
\State \Comment{Phase 3: Enhancement Generation}
\State $\mathbb{E} \leftarrow \emptyset$
\For{$G_j \in \mathcal{G}$}
    \State $(E_j^{(c)}, E_j^{(s)}, E_j^{(r)}) \leftarrow \textsc{Synthesize}(\mathcal{M}_\phi, G_j)$
    \State $\mathbb{E} \leftarrow \mathbb{E} \cup \{(E_j^{(c)}, E_j^{(s)}, E_j^{(r)})\}$
\EndFor

\Statex
\State \Comment{Aggregate into final prompts}
\State $(P'^{(c)}, P'^{(s)}, P'^{(r)}) \leftarrow \textsc{Aggregate}(P, \mathbb{E}, \mathcal{G})$

\Statex
\State \Return $(P'^{(c)}, P'^{(s)}, P'^{(r)})$

\end{algorithmic}
\end{algorithm}

Table~\ref{tab:notation} summarizes the key notation used in the algorithms.

\begin{table}[h]
\centering
\small
\begin{tabular}{cl}
\hline
\textbf{Symbol} & \textbf{Description} \\
\hline
$\mathcal{Q}$ & Failed question set \\
$\mathbb{A}$ & Set of failure analyses \\
$\mathcal{G}$ & Partition of clustered failures \\
$\mathcal{M}_\phi$ & Analyzer model \\
$\mathcal{E}$ & Error taxonomy \\
\hline
$\tau$ & Question type \\
$\mathcal{T}$ & Topic set \\
$\epsilon$ & Error type \\
$\rho$ & Root cause \\
$\mu$ & Specific mistake \\
\hline
$E^{(c)}$ & Concise enhancement \\
$E^{(s)}$ & Specific enhancement \\
$E^{(r)}$ & Reasoning enhancement \\
$\oplus$ & Prompt concatenation \\
\hline
\end{tabular}
\caption{Notation used in MARS algorithms.}
\label{tab:notation}
\end{table}

\section{Enhancement Variants}
\label{appendix:enhancement_variants}

Our method generates three enhancement variants for each type-topic group, each serving a distinct purpose during inference. Table~\ref{tab:enhancement_variants} summarizes their characteristics.

\begin{table*}[h]
\centering
\resizebox{0.7\linewidth}{!}{
\begin{tabular}{llll}
\toprule
\textbf{Variant} & \textbf{Purpose} & \textbf{Content} & \textbf{Length} \\
\midrule
Concise ($E^{(c)}$) & Quick reference & Warnings and key points & Short \\
Specific ($E^{(s)}$) & Thorough guidance & Mistakes, verification steps, approach & Detailed \\
Reasoning ($E^{(r)}$) & Self-correction & Minimal hints for discovery & Minimal \\
\bottomrule
\end{tabular}%
}
\caption{Comparison of three enhancement variants.}
\label{tab:enhancement_variants}
\end{table*}

The \textbf{concise} variant $E^{(c)}$ provides brief warnings derived from common mistakes, suitable for scenarios where inference cost is a concern. The \textbf{specific} variant $E^{(s)}$ includes detailed verification steps and explicit reasoning strategies, appropriate when accuracy is prioritized over efficiency. The \textbf{reasoning} variant $E^{(r)}$ offers minimal guidance designed to trigger self-correction without over-constraining the model's reasoning process.

\section{Prompts}
\label{sec:appendix-prompts}

This appendix provides the complete prompt templates used in our experiments. All prompts use a structured XML-style output format with \texttt{<reasoning>} and \texttt{<answer>} tags to facilitate consistent response parsing. The placeholder \texttt{\{question\}} is replaced with the specific problem instance at inference time.

\subsection{Zero-Shot Prompting}
\label{sec:zero-shot}

The zero-shot baseline provides the question directly without any demonstrations or reasoning instructions.

\begin{figure*}[h]
\centering
\includegraphics[width=0.85\linewidth]{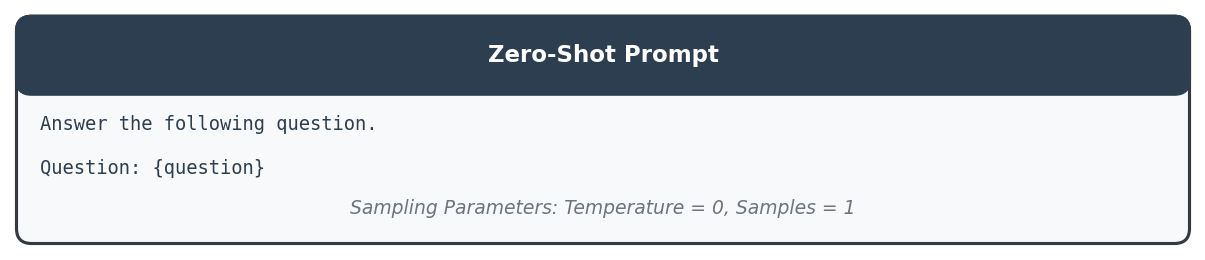}
\caption{Zero-shot prompt template.}
\label{fig:prompt-zero-shot}
\end{figure*}

\subsection{Zero-Shot Chain-of-Thought}
\label{sec:zero-shot-cot}

Zero-shot chain-of-thought prompting \citep{kojima2022large} elicits step-by-step reasoning without providing exemplars.

\begin{figure*}[h]
\centering
\includegraphics[width=0.85\linewidth]{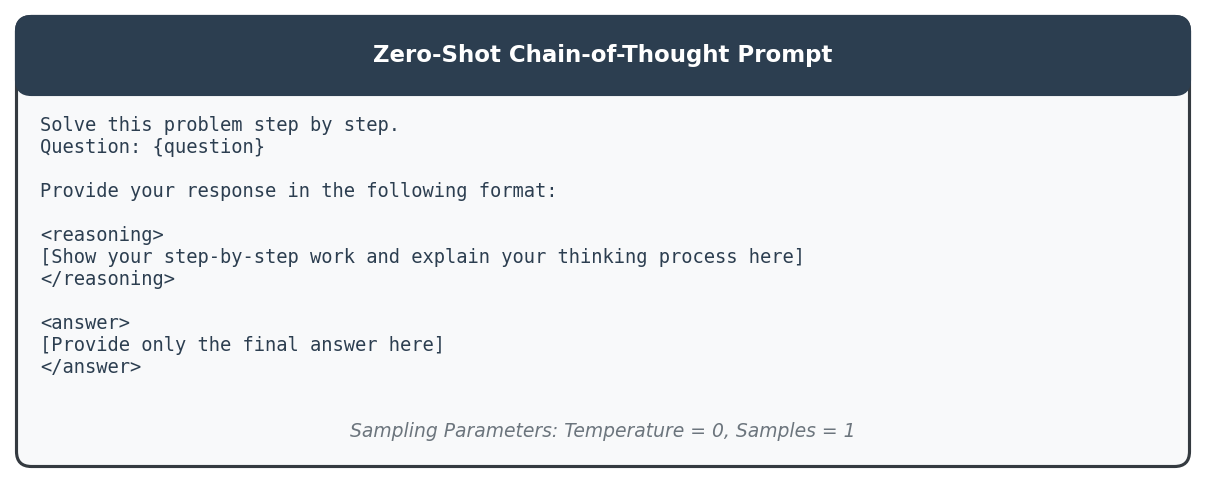}
\caption{Zero-shot chain-of-thought prompt template.}
\label{fig:prompt-zero-shot-cot}
\end{figure*}

\subsection{Few-Shot Chain-of-Thought}
\label{sec:few-shot-cot}

Few-shot chain-of-thought prompting \citep{wei2022chain} provides exemplars demonstrating the reasoning process.

\begin{figure*}[h]
\centering
\includegraphics[width=0.85\linewidth]{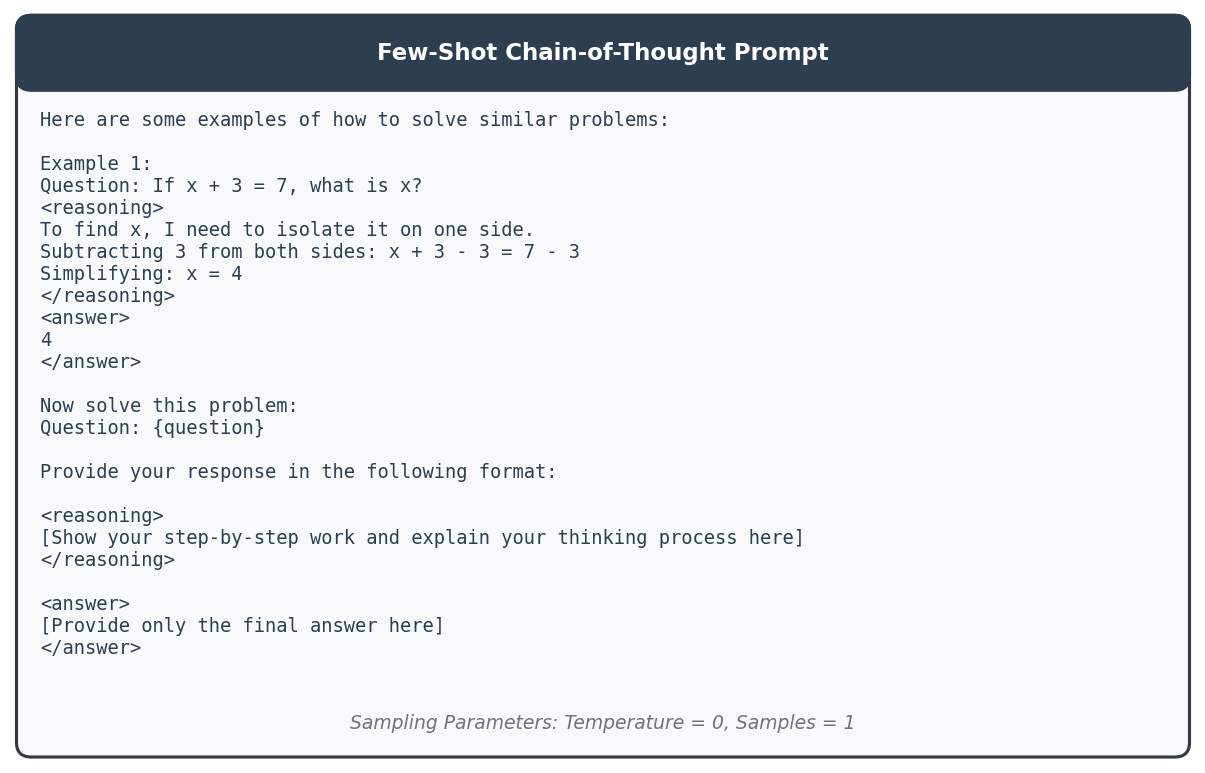}
\caption{Few-shot chain-of-thought prompt template with one demonstration example.}
\label{fig:prompt-few-shot-cot}
\end{figure*}

\subsection{Self-Consistency}
\label{sec:self-consistency}

Self-consistency \citep{wang2023selfconsistency} samples multiple reasoning paths and aggregates answers via majority voting. We evaluate configuration with 10 samples.

\begin{figure*}[h]
\centering
\includegraphics[width=0.85\linewidth]{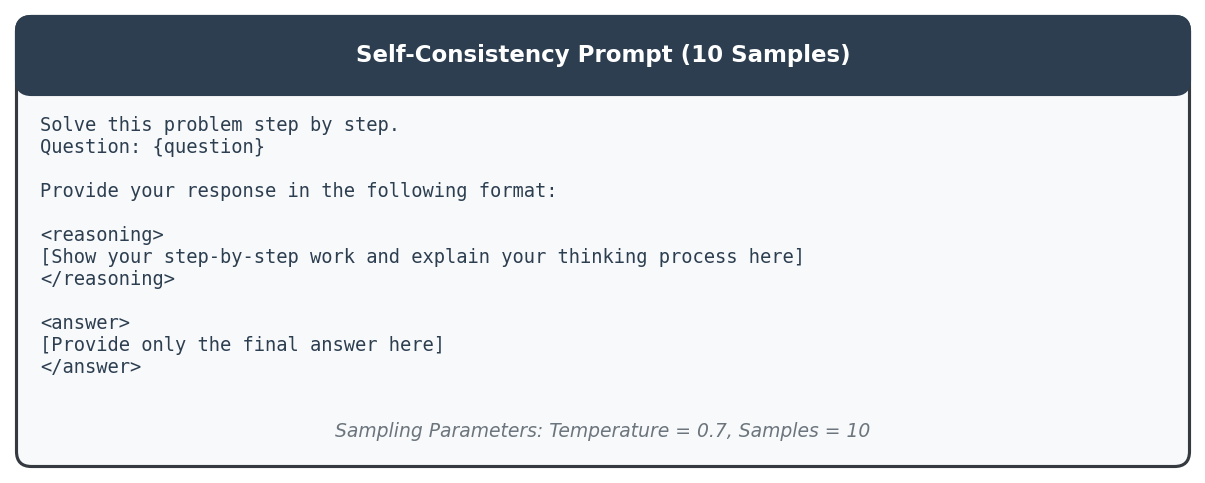}
\caption{Self-consistency prompt template (10 samples).}
\label{fig:prompt-sc10}
\end{figure*}

\subsection{Self-Refine}
\label{sec:self-refine}

Self-refine \citep{madaan2023selfrefine} enables iterative improvement of responses through self-feedback.

\begin{figure*}[h]
\centering
\includegraphics[width=0.85\linewidth]{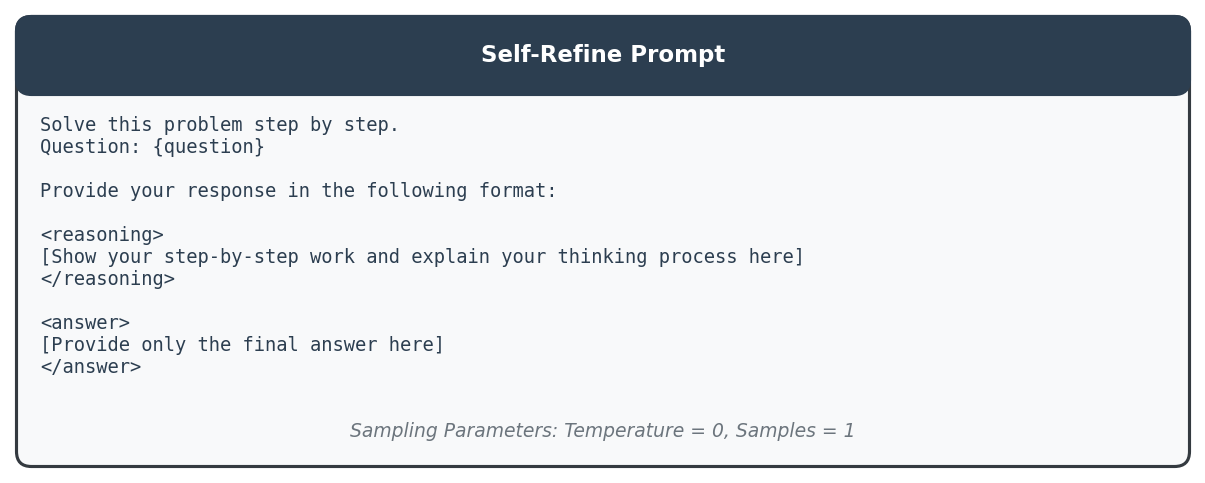}
\caption{Self-refine prompt template.}
\label{fig:prompt-self-refine}
\end{figure*}

\subsection{Enhancement Analyzer}
\label{sec:enhancement-analyzer}

The Enhancement Analyzer is the first LLM-powered agent in our zero-shot enhancement pipeline. It performs individual failure analysis by examining each incorrectly answered GPQA question to determine the precise cause of error. Given a failed question along with the model's predicted answer and reasoning, the analyzer classifies the question type (factual, conceptual, calculation, application, analysis, or comparison), identifies specific scientific topics, and diagnoses the error type (e.g., conceptual misunderstanding, calculation error, misreading, incomplete analysis, wrong elimination, or knowledge gap). The agent produces a structured JSON output containing the root cause explanation, the specific reasoning step that failed, required domain knowledge, and factors contributing to the question's difficulty. This granular analysis enables downstream pattern recognition across multiple failures.

\begin{figure*}[h]
\centering
\includegraphics[width=0.85\linewidth]{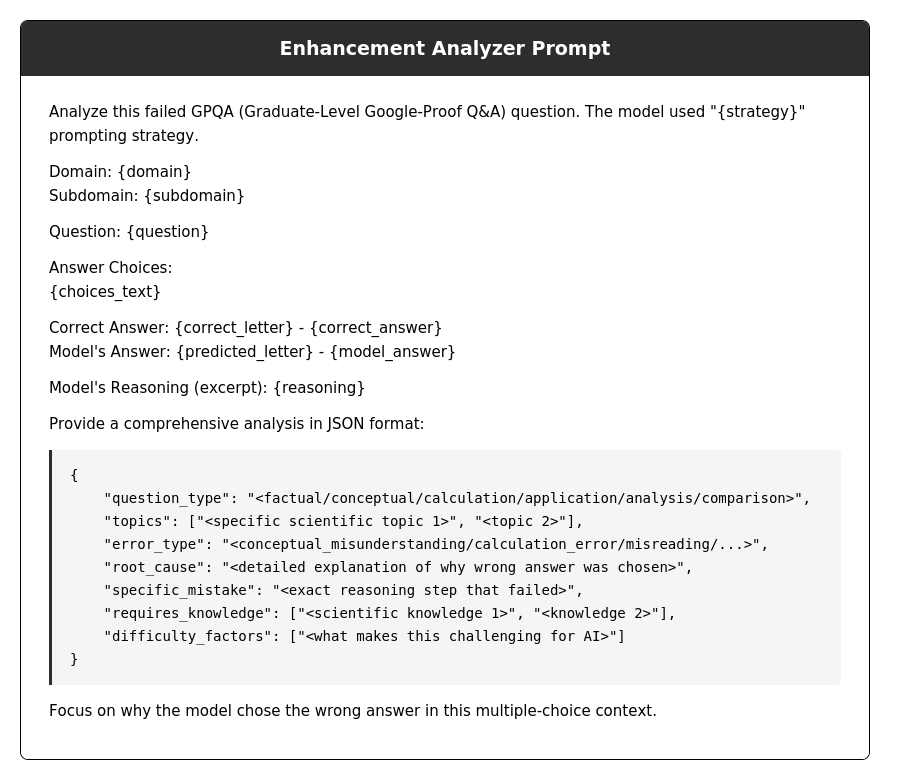}
\caption{Enhancement Analyzer prompt template for individual failure analysis.}
\label{fig:prompt-enhancement-analyzer}
\end{figure*}

\subsection{Enhancement Synthesizer}
\label{sec:enhancement-synthesizer}

The Enhancement Synthesizer is the second LLM-powered agent that operates on grouped failures sharing common characteristics. After the Enhancement Analyzer processes individual questions, failures are clustered by question type and topic. The Synthesizer then analyzes each cluster to identify recurring error patterns, synthesize shared root causes, and generate targeted enhancement strategies. For each type-topic group, it produces common mistake patterns, critical warnings, verification steps, topic-specific guidance, and a concise prompt addition designed to prevent similar errors. This synthesized knowledge is then used to construct three variants of enhanced prompts (concise, specific, and reasoning-focused), each tailored to address the identified weaknesses while maintaining the base prompting strategy's structure.

\begin{figure*}[h]
\centering
\includegraphics[width=0.85\linewidth]{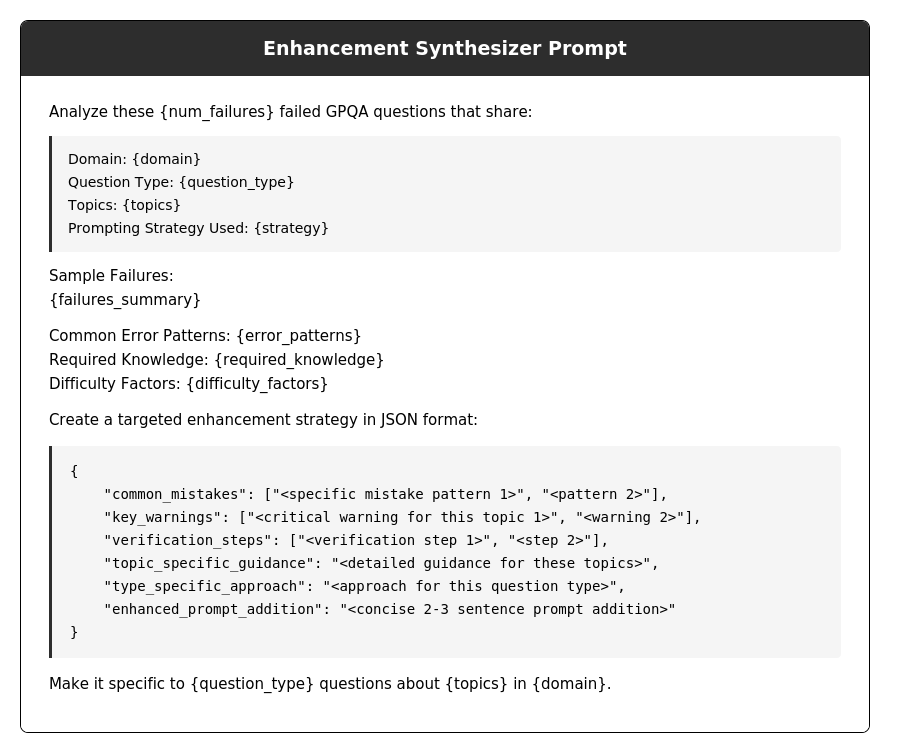}
\caption{Enhancement Synthesizer prompt template for pattern-based enhancement generation.}
\label{fig:prompt-enhancement-synthesizer}
\end{figure*}
\subsection{Model Configuration}
\label{sec:model-config}

Table~\ref{tab:model-config} summarizes the model configuration used across all experiments.

\begin{table}[h]
\centering
\small
\begin{tabular}{ll}
\toprule
\textbf{Parameter} & \textbf{Value} \\
\midrule
Default Model & GPT-4o \\
Evaluation Model & o3-mini \\
Max Tokens & 2000 \\
Timeout & 30 seconds \\
\bottomrule
\end{tabular}
\caption{Model configuration settings.}
\label{tab:model-config}
\end{table}

\begin{table}[h]
\centering
\resizebox{\linewidth}{!}{
\begin{tabular}{lccc}
\toprule
\textbf{Component} & \textbf{Concise} & \textbf{Reasoning} & \textbf{Specific} \\
\midrule
Explicit warnings & Yes & No & Yes \\
Action sequences & Yes & No & Yes \\
Process guidance & Brief & Detailed & Detailed \\
Verification steps & No & Yes & Yes \\
Approach description & No & Yes & Yes \\
\midrule
Relative length & Short & Medium & Long \\
\bottomrule
\end{tabular}}
\caption{Structural comparison of MARS enhancement types.}
\label{tab:enhancement_comparison}
\end{table}
\section{Enhancement Generation with Open-source Model}
\label{appendix:qwen_results}
\begin{table*}[h]
\centering
\resizebox{0.6\linewidth}{!}{
\begin{tabular}{lccccc}
\toprule
\textbf{Method} & \textbf{Baseline} & \textbf{Concise} & \textbf{Reasoning} & \textbf{R+C} & \textbf{Hybrid} \\
\midrule
Zero-shot & 11.82 & 12.73 & 11.82 & 13.64 & 19.10 \\
Zero-shot-CoT & 16.40 & 18.18 & 15.45 & 13.64 & 22.70 \\
Self-refine & 36.36 & 37.27 & 39.09 & 37.27 & 48.20 \\
Self-consistency & 18.20 & 19.09 & 17.27 & 22.73 & 32.60 \\
\bottomrule
\end{tabular}}
\caption{GPQA performance (\%) with Qwen-generated enhancements.}
\label{tab:qwen_gpqa}
\end{table*}

\begin{table*}[h]
\centering
\resizebox{0.6\linewidth}{!}{
\begin{tabular}{lccccc}
\toprule
\textbf{Method} & \textbf{Baseline} & \textbf{Concise} & \textbf{Reasoning} & \textbf{R+C} & \textbf{Hybrid} \\
\midrule
Zero-shot & 11.82 & 12.73 & 11.82 & 13.64 & 19.10 \\
Zero-shot-CoT & 16.40 & 18.18 & 15.45 & 13.64 & 22.70 \\
Self-refine & 36.36 & 37.27 & 39.09 & 37.27 & 48.20 \\
Self-consistency & 18.20 & 19.09 & 17.27 & 22.73 & 32.60 \\
\bottomrule
\end{tabular}}
\caption{OMNI-math performance (\%) with Qwen-generated enhancements.}
\label{tab:qwen_omni}
\end{table*}

To evaluate the generalizability of MARS enhancements across different enhancement generators, we replicate experiments using Qwen2.5-72B-Instruct-Turbo~\cite{qwen2.5} for enhancement generation on one knowledge coverage dataset (GPQA) and one reasoning capacity dataset (OMNI-math). Results are presented in Tables~\ref{tab:qwen_gpqa} and~\ref{tab:qwen_omni}.

\paragraph{Analysis.} Qwen-generated enhancements demonstrate consistent improvement patterns across both datasets. The hybrid enhancement achieves the highest gains, with self-refine + hybrid reaching 48.20\% on GPQA (a 32.6\% relative improvement over baseline). Individual enhancements show modest gains: concise enhancement improves zero-shot by 7.7\% relative, while reasoning+concise provides the largest single-enhancement gain for self-consistency (24.9\% relative). These results confirm that MARS enhancement effectiveness generalizes across enhancement generators, though optimal enhancement selection remains task-dependent.

\section{One-Turn MARS Enhancement Samples}
\label{appendix:enhancement_types}

MARS generates three types of one-turn prompt enhancements from the same error analysis, each with different structures and verbosity levels. We demonstrate each type using the Algebra\_Equations category under zero-shot prompting.

\subsection{Concise Enhancement}

\begin{figure*}[htbp]
    \centering
    \includegraphics[width=0.95\textwidth]{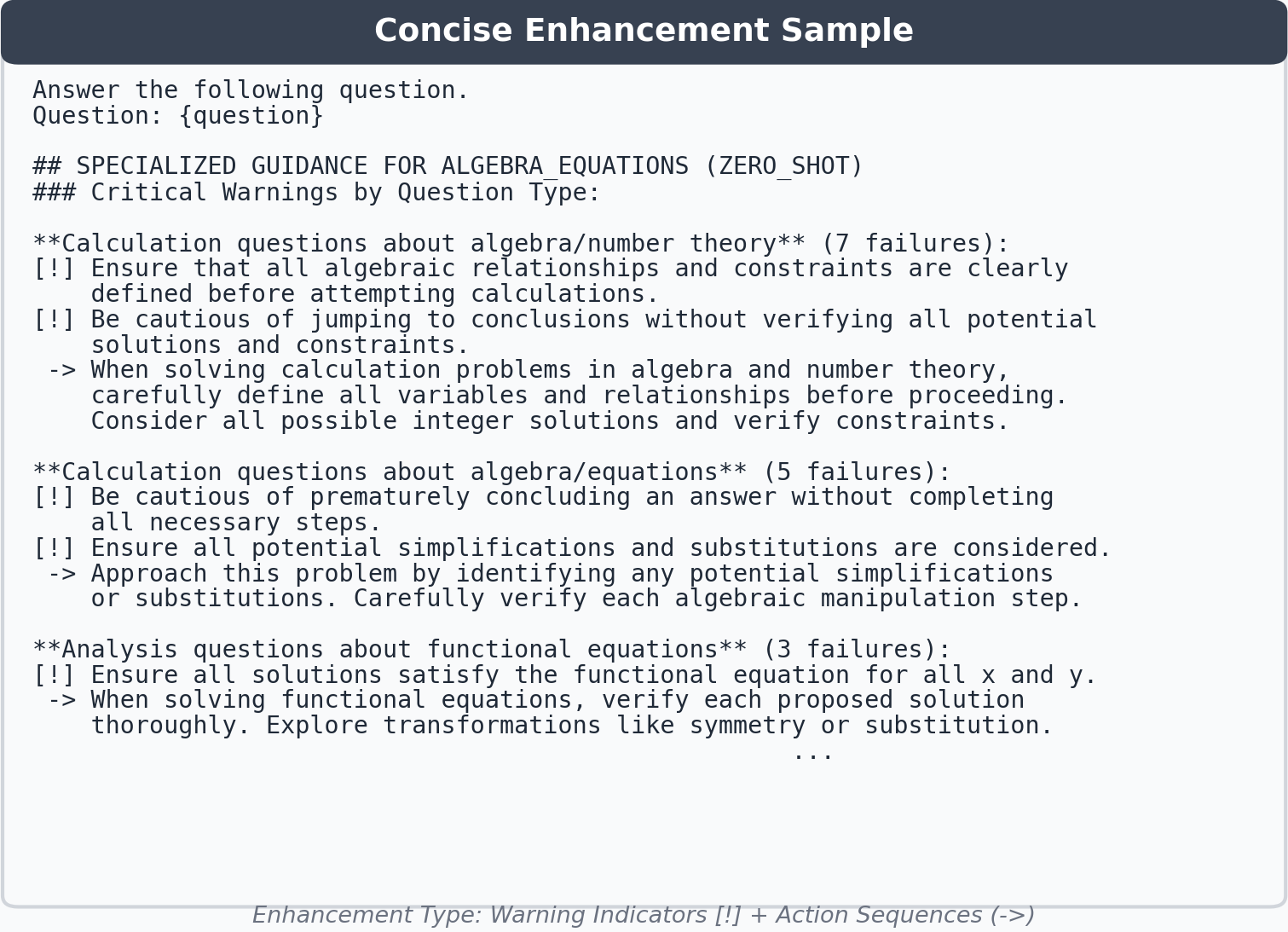}
    \caption{Concise enhancement template structure. Warnings identify failure patterns; arrows provide actionable guidance.}
    \label{fig:template_concise}
\end{figure*}

The concise enhancement (Figure~\ref{fig:template_concise}) provides compact, action-oriented guidance organized by type-topic groups. Each group includes warning indicators (\texttt{[!]}) highlighting common failure patterns with failure counts, followed by action arrows (\texttt{->}) specifying recommended problem-solving procedures. This format minimizes prompt length while preserving critical guidance.

\subsection{Reasoning Enhancement}

\begin{figure*}[htbp]
    \centering
    \includegraphics[width=0.95\textwidth]{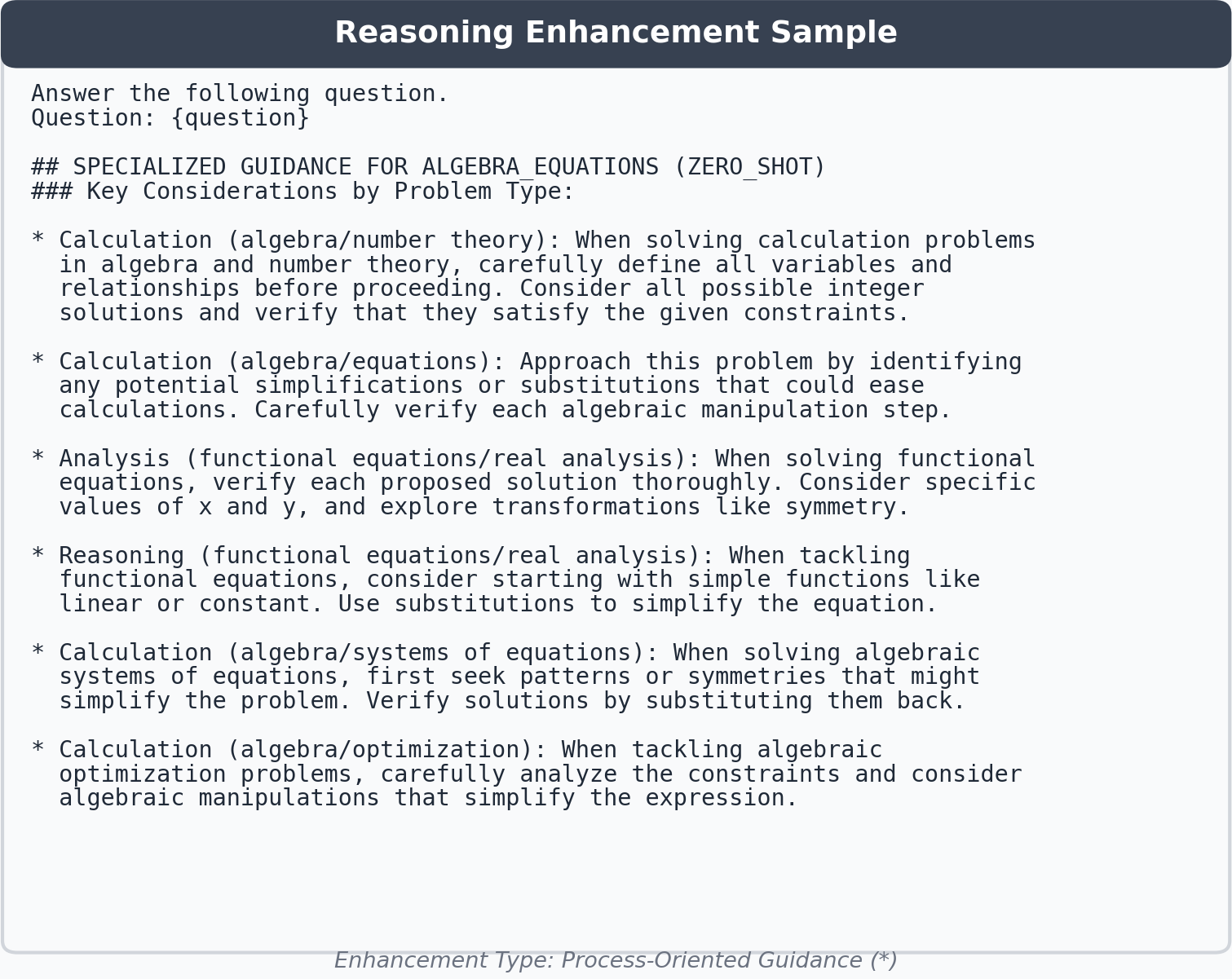}
    \caption{Reasoning enhancement template structure. Bullet points provide process-oriented guidance for each problem type.}
    \label{fig:template_reasoning}
\end{figure*}

The reasoning enhancement (Figure~\ref{fig:template_reasoning}) emphasizes problem-solving strategies over explicit warnings. Each type-topic group receives a bulleted consideration (\texttt{*}) describing the recommended reasoning approach. This format focuses on \textit{how} to think about problems rather than \textit{what} to avoid, making it effective for tasks requiring flexible reasoning.

\subsection{Specific Enhancement (Reasoning + Concise)}

\begin{figure*}[htbp]
    \centering
    \includegraphics[width=0.95\textwidth]{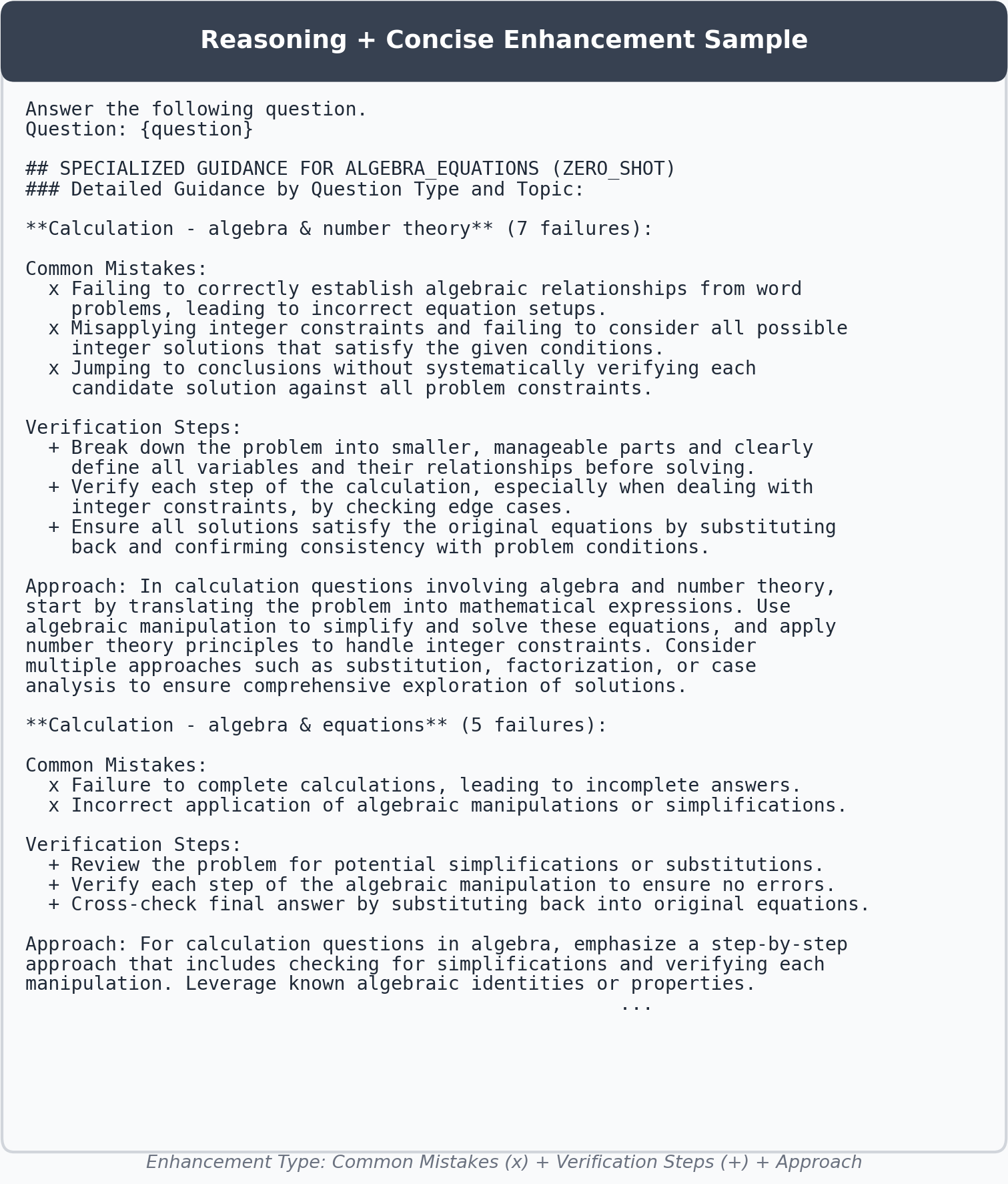}
    \caption{Specific enhancement template structure. Combines explicit mistake warnings, verification steps, and methodological guidance.}
    \label{fig:template_specific}
\end{figure*}

The specific enhancement (Figure~\ref{fig:template_specific}) combines both approaches into a comprehensive three-part structure for each type-topic group: (1) \textbf{Common Mistakes} (\texttt{x}) explicitly enumerate failure patterns; (2) \textbf{Verification Steps} (\texttt{+}) provide concrete validation actions; (3) \textbf{Approach} describes the recommended problem-solving methodology. This format maximizes guidance completeness at the cost of increased prompt length.

\subsection{Enhancement Comparison}

Table~\ref{tab:enhancement_comparison} summarizes the structural differences between enhancement types.

The hybrid enhancement strategy evaluates all three types on a validation set and selects the optimal enhancement for each question, combining their complementary strengths.

\section{Computational Cost Analysis}
\label{appendix:computational_cost}

We compare the computational costs of MARS against recursive self-improvement baselines: Gödel Agent~\cite{yin2025godelagent} and Meta Agent Search (ADAS)~\cite{hu2025adas}. Costs are estimated based on reported experimental configurations and current API pricing.

\subsection{Baseline Method Costs}

\paragraph{Meta Agent Search.} As reported in~\cite{hu2025adas}, Meta Agent Search runs for \textbf{25 iterations}, where each iteration involves: (1) the meta agent (GPT-4) programming a new agent design, (2) self-reflection refinement (2 iterations per proposal, plus up to 3 error-correction refinements), and (3) evaluation on validation data using GPT-3.5. The authors report a total cost of approximately \textbf{\$300} across four benchmarks (DROP, MGSM, MMLU, GPQA). This high cost stems from the extensive GPT-4 usage for iterative agent design and the growing archive of discovered agents that must be included in each subsequent prompt.

\paragraph{Gödel Agent.} As reported in~\cite{yin2025godelagent}, Gödel Agent performs \textbf{30 recursive self-improvements} across four benchmarks, with a total cost of approximately \textbf{\$15}. The framework uses GPT-4o for self-modification and GPT-3.5 for policy evaluation. The reduced cost compared to Meta Agent Search is attributed to continuous self-optimization that enables faster convergence. However, the authors note that the main cost driver is the continuously growing historical memory, suggesting that longer optimization runs would incur substantially higher costs.

\subsection{MARS Cost Estimation}

MARS operates in a \textbf{single recurrence cycle} with three phases:

\begin{enumerate}
    \item \textbf{Evaluation Phase}: The analyzer model processes each failed question to produce structured diagnoses. For a typical benchmark with $\sim$200 failed questions, this requires $\sim$200 API calls.
    
    \item \textbf{Failure Allocation Phase}: Pure computational grouping with no API calls required.
    
    \item \textbf{Enhancement Generation Phase}: The synthesizer generates enhancements for each type-topic group. With $\sim$15--20 groups per benchmark, this requires $\sim$40--60 API calls (including concise, reasoning, and specific variants).
\end{enumerate}

Using GPT-3.5-turbo (\$0.0005/1K input, \$0.0015/1K output tokens) for both analysis and synthesis, the estimated cost per benchmark is:

\begin{table}[h]
\centering
\resizebox{\linewidth}{!}{
\begin{tabular}{lrrr}
\toprule
\textbf{Phase} & \textbf{API Calls} & \textbf{Tokens (K)} & \textbf{Est. Cost} \\
\midrule
Evaluation & $\sim$200 & $\sim$400 & $\sim$\$0.40 \\
Enhancement Gen. & $\sim$50 & $\sim$150 & $\sim$\$0.15 \\
\midrule
\textbf{Total (per benchmark)} & $\sim$250 & $\sim$550 & $\sim$\$0.55 \\
\textbf{Total (4 benchmarks)} & $\sim$1,000 & $\sim$2,200 & $\sim$\$2.20 \\
\bottomrule
\end{tabular}}
\caption{Estimated MARS computational cost using GPT-3.5-turbo.}
\label{tab:mars_cost}
\end{table}

\subsection{Cost Comparison Summary}

Table~\ref{tab:cost_comparison} summarizes the computational requirements across methods.

\begin{table*}[h]
\centering
\resizebox{0.65\linewidth}{!}{
\begin{tabular}{lccccc}
\toprule
\textbf{Method} & \textbf{Iterations} & \textbf{Meta Model} & \textbf{Eval Model} & \textbf{Cost (4 benchmarks)} & \textbf{Relative} \\
\midrule
Meta Agent Search & 25 & GPT-4 & GPT-3.5 & $\sim$\$300 & 136$\times$ \\
Gödel Agent & 30 & GPT-4o & GPT-3.5 & $\sim$\$15 & 6.8$\times$ \\
\textbf{MARS (Ours)} & \textbf{1} & GPT-3.5 & GPT-3.5 & $\sim$\$2.20 & \textbf{1$\times$} \\
\bottomrule
\end{tabular}}
\caption{Computational cost comparison across self-improvement methods. Costs based on reported values~\cite{yin2025godelagent,hu2025adas} and our estimates.}
\label{tab:cost_comparison}
\end{table*}

\subsection{Analysis}

The cost reduction achieved by MARS stems from three design decisions:

\begin{enumerate}
    \item \textbf{Single-cycle learning}: While recursive methods require 25--30 iterations to converge, MARS consolidates learning into one pass, eliminating the multiplicative cost of iteration.
    
    \item \textbf{No growing context}: Recursive methods accumulate historical memory (Gödel Agent) or an archive of discovered agents (Meta Agent Search), causing token usage to grow with each iteration. MARS processes fixed-size inputs throughout.
    
    \item \textbf{Efficient model selection}: MARS uses GPT-3.5 for both analysis and synthesis, while recursive methods require GPT-4/GPT-4o for meta-level reasoning. As shown in Appendix~\ref{appendix:qwen_results}, enhancement quality is robust to generator model choice.
\end{enumerate}

The cost-performance trade-off favors MARS: at 136$\times$ lower cost than Meta Agent Search and 6.8$\times$ lower than Gödel Agent, MARS achieves comparable or superior performance (Table~\ref{tab:main_results}). For applications requiring maximum performance regardless of cost, MARS can be applied iteratively---using enhanced prompts to generate new failure sets for further refinement---while still maintaining substantial cost advantages over recursive baselines.

\section{MARS Implementation Details}
\label{appendix:code}

This appendix provides key code snippets from the MARS implementation.

\subsection{Data Structures}

Listing~\ref{lst:datastructures} defines the core data structures corresponding to Equations~\ref{eq:analysis}--\ref{eq:error_profile}.

\begin{lstlisting}[language=Python, caption={Core data structures for MARS pipeline.}, label={lst:datastructures}]
@dataclass
class IndividualFailureAnalysis:
    """Structured analysis A_i = (tau_i, T_i, epsilon_i, rho_i, mu_i)"""
    question_id: str
    question_text: str
    question_type: str      # tau_i in Y (question type)
    topics: List[str]       # T_i subset of D (topic set)
    error_type: str         # epsilon_i in E (error taxonomy)
    root_cause: str         # rho_i (reasoning deficit)
    specific_mistake: str   # mu_i (divergence point)
    requires_knowledge: List[str]
    difficulty_factors: List[str]

@dataclass
class QuestionTypeTopicGroup:
    """Group G_j with error profile Psi_j = (E_j, R_j, F_j)"""
    question_type: str
    topics: List[str]
    failures: List[IndividualFailureAnalysis]
    common_error_patterns: List[str]  # E_j
    shared_root_causes: List[str]     # R_j
    required_knowledge: Set[str]
    key_difficulty_factors: List[str] # F_j

@dataclass
class TypeTopicEnhancement:
    """Enhancement variants for type-topic group:
       E^(c): concise, E^(r): reasoning, E^(c+r): specific"""
    question_type: str
    topics: List[str]
    num_questions: int              # |G_j| for weight w_j
    # For concise enhancement E^(c):
    key_warnings: List[str]         # Warning indicators [!]
    # For reasoning enhancement E^(r):
    enhanced_prompt_addition: str   # Process-oriented guidance
    # For specific enhancement E^(c+r) = concise + reasoning:
    common_mistakes: List[str]      # Explicit error patterns (x)
    verification_steps: List[str]   # Validation actions (+)
    type_specific_approach: str     # Methodological guidance
\end{lstlisting}

\subsection{Phase 1: Individual Failure Analysis}

Listing~\ref{lst:individual_analysis} shows the evaluation phase that produces structured analyses $\mathcal{A}_i$ for each failed question.

\begin{lstlisting}[language=Python, caption={Code for individual failure analysis (Evaluation phase).}, label={lst:individual_analysis}]
def analyze_individual_failure(self, failure: Dict, 
                               strategy: str) -> IndividualFailureAnalysis:
    question = failure.get('question', '')
    correct_answer = failure.get('correct_answer', '')
    model_answer = failure.get('predicted_answer', '')
    
    prompt = f"""Analyze this failed question using "{strategy}" strategy.
Question: {question[:2000]}
Correct: {correct_answer[:500]}
Model Answer: {model_answer[:500]}

Provide JSON analysis:
{{
    "question_type": "<factual/conceptual/calculation/application>",
    "topics": ["<topic_1>", "<topic_2>"],
    "error_type": "<conceptual_misunderstanding/calculation_error/...>",
    "root_cause": "<fundamental reasoning deficit>",
    "specific_mistake": "<exact step where logic diverged>",
    "requires_knowledge": ["<knowledge_1>"],
    "difficulty_factors": ["<factor_1>"]
}}"""
    
    result = call_llm(self.client, self.model,
                      [{"role": "user", "content": prompt}],
                      temperature=0.3, max_tokens=800)
    data = json.loads(extract_json(result))
    
    return IndividualFailureAnalysis(
        question_type=data.get('question_type'),
        topics=data.get('topics', []),
        error_type=data.get('error_type'),
        root_cause=data.get('root_cause', ''),
        specific_mistake=data.get('specific_mistake', ''),
        requires_knowledge=data.get('requires_knowledge', []),
        difficulty_factors=data.get('difficulty_factors', []))
\end{lstlisting}

\subsection{Phase 2: Type-Topic Grouping}

Listing~\ref{lst:grouping} implements the grouping function $\kappa$ (Equation~\ref{eq:grouping}) that partitions analyses into groups $\mathcal{G} = \{G_j\}$.

\begin{lstlisting}[language=Python, caption={Code for failure allocation via type-topic grouping.}, label={lst:grouping}]
def group_by_type_topic(self, analyses: List[IndividualFailureAnalysis]
                        ) -> List[QuestionTypeTopicGroup]:
    """Apply grouping function kappa: A -> Y x 2^D"""
    
    # Partition by composite key (tau_i, T_i)
    groups = defaultdict(list)
    for analysis in analyses:
        key = (analysis.question_type, 
               frozenset(analysis.topics[:2]))
        groups[key].append(analysis)
    
    type_topic_groups = []
    for (q_type, topics), group_analyses in groups.items():
        # Aggregate into error profile Psi_j
        error_patterns = [a.error_type for a in group_analyses]
        root_causes = [a.root_cause for a in group_analyses]
        required_knowledge = set()
        difficulty_factors = []
        
        for a in group_analyses:
            required_knowledge.update(a.requires_knowledge)
            difficulty_factors.extend(a.difficulty_factors)
        
        group = QuestionTypeTopicGroup(
            question_type=q_type,
            topics=list(topics),
            failures=group_analyses,
            common_error_patterns=list(set(error_patterns)),
            shared_root_causes=list(set(root_causes)),
            required_knowledge=required_knowledge,
            key_difficulty_factors=list(set(difficulty_factors)))
        type_topic_groups.append(group)
    
    return type_topic_groups
\end{lstlisting}

\subsection{Phase 3: Enhancement Generation}

Listing~\ref{lst:enhancement_gen} implements the enhancement function $\xi$ (Equation~\ref{eq:enhancement_function}) and prompt aggregation (Equation~\ref{eq:final_prompt}). The three enhancement variants are: (1) \textbf{concise} $E^{(c)}$: warning indicators + action sequences; (2) \textbf{reasoning} $E^{(r)}$: process-oriented guidance; and (3) \textbf{specific} $E^{(c+r)}$: the combination of concise and reasoning, providing common mistakes, verification steps, and methodological approach.

\begin{lstlisting}[language=Python, caption={Code for enhancement generation and prompt aggregation.}, label={lst:enhancement_gen}]
def create_enhanced_prompts(self, base_prompt: str,
                            enhancements: List[TypeTopicEnhancement],
                            strategy: str, category: str) -> Dict[str, str]:
    """Generate P'^(c) and P'^(r) via weighted aggregation"""
    
    # Sort by |G_j| for weight w_j (Eq. 5)
    sorted_enh = sorted(enhancements, 
                        key=lambda e: e.num_questions, reverse=True)
    all_prompts = {}
    
    # Concise enhancement E^(c): warnings + action sequences
    if 'concise' in self.enhancement_types:
        text = f"\n## GUIDANCE FOR {category.upper()}\n"
        text += "### Critical Warnings by Question Type:\n\n"
        for enh in sorted_enh[:8]:  # Top-weighted groups
            text += f"**{enh.question_type} ({'/'.join(enh.topics)})** "
            text += f"({enh.num_questions} failures):\n"
            text += "[!] " + " | ".join(enh.key_warnings[:3]) + "\n"
            text += f" -> {enh.enhanced_prompt_addition}\n\n"
        all_prompts['concise'] = base_prompt + text  # P' = P + E
    
    # Reasoning enhancement E^(r): process-oriented hints
    if 'reasoning' in self.enhancement_types:
        text = f"\n## GUIDANCE FOR {category.upper()}\n"
        text += "### Key Considerations by Problem Type:\n\n"
        for enh in sorted_enh[:6]:
            text += f"* {enh.question_type} ({'/'.join(enh.topics)}): "
            text += f"{enh.enhanced_prompt_addition}\n"
        all_prompts['reasoning'] = base_prompt + text
    
    # Specific enhancement E^(c+r): combines concise + reasoning
    # Includes: mistakes (concise) + verification (concise) + approach (reasoning)
    if 'specific' in self.enhancement_types:
        text = f"\n## GUIDANCE FOR {category.upper()}\n"
        for enh in sorted_enh[:10]:
            text += f"**{enh.question_type} - {' & '.join(enh.topics)}**\n"
            # From concise: explicit warnings as mistake patterns
            text += "Common Mistakes:\n"
            for m in enh.common_mistakes[:3]: text += f"  x {m}\n"
            # From concise: action sequences as verification
            text += "Verification Steps:\n"
            for s in enh.verification_steps[:4]: text += f"  + {s}\n"
            # From reasoning: process-oriented approach
            text += f"Approach: {enh.type_specific_approach}\n\n"
        all_prompts['specific'] = base_prompt + text
    
    return all_prompts
\end{lstlisting}

\subsection{Hybrid Selection}

Listing~\ref{lst:hybrid} implements the hybrid strategy (Equation~\ref{eq:hybrid_selection}) that selects optimal enhancement per category.

\begin{lstlisting}[language=Python, caption={Code for hybrid enhancement selection.}, label={lst:hybrid}]
def select_hybrid_enhancement(self, val_data: Dict[str, List],
                              enhancements: Dict) -> Dict[str, str]:
    """Select E*_c = argmax Acc(E, V_c) for each category c
       where E in {E^(c), E^(r), E^(c+r)} (concise, reasoning, specific)"""
    
    optimal = {}
    # specific = concise + reasoning (c+r)
    etypes = ['concise', 'reasoning', 'specific']
    
    for category, val_questions in val_data.items():
        best_acc, best_type = 0, 'concise'
        
        for etype in etypes:
            enhanced_prompt = enhancements.get(f"{category}_{etype}")
            if not enhanced_prompt: continue
            
            correct = sum(1 for q in val_questions
                         if self.evaluate(enhanced_prompt, q) 
                            == q['correct_answer'])
            accuracy = correct / len(val_questions)
            
            if accuracy > best_acc:
                best_acc, best_type = accuracy, etype
        
        optimal[category] = best_type
        print(f"  {category}: '{best_type}' (acc: {best_acc:.1%})")
    
    return optimal
\end{lstlisting}
\end{document}